\documentclass[10pt]{article}

\usepackage[preprint]{templates/tmlr/tmlr}

\usepackage{amsmath}
\usepackage{amssymb}

\usepackage{multirow}
\usepackage[table]{xcolor}
\usepackage{xspace}
\usepackage{chemfig}
\usepackage{subcaption}
\usepackage{threeparttable}

\newcommand{\thename}{CDI\xspace}

\usepackage{hyperref}
\usepackage{url}

\title{Parameter Inference and Uncertainty Quantification with Diffusion Models:\\ Extending CDI to 2D Spatial Conditioning}

\author{\name Dmitrii Torbunov \email dtorbunov@bnl.gov \\
       \addr Brookhaven National Laboratory, USA\\
       \AND
       \name Yihui Ren \email yren@bnl.gov \\
       \addr Brookhaven National Laboratory, USA\\
       \AND
       \name Lijun Wu \email ljwu@bnl.gov \\
       \addr Brookhaven National Laboratory, USA\\
       \AND
       \name Yimei Zhu \email zhu@bnl.gov \\
       \addr Brookhaven National Laboratory, USA\\
}

\begin{document}

\maketitle

\begin{abstract}
Uncertainty quantification is critical in scientific inverse problems to distinguish identifiable parameters from those that remain ambiguous given available measurements.
The Conditional Diffusion Model-based Inverse Problem Solver (CDI) has previously demonstrated effective probabilistic inference for one-dimensional temporal signals, but its applicability to higher-dimensional spatial data remains unexplored.
We extend CDI to two-dimensional spatial conditioning, enabling probabilistic parameter inference directly from spatial observations.
We validate this extension on convergent beam electron diffraction (CBED) parameter inference -- a challenging multi-parameter inverse problem in materials characterization where sample geometry, electronic structure, and thermal properties must be extracted from 2D diffraction patterns.
Using simulated CBED data with ground-truth parameters, we demonstrate that CDI produces well-calibrated posterior distributions that accurately reflect measurement constraints: tight distributions for well-determined quantities and appropriately broad distributions for ambiguous parameters.
In contrast, standard regression methods—while appearing accurate on aggregate metrics—mask this underlying uncertainty by predicting training set means for poorly constrained parameters.
Our results confirm that CDI successfully extends from temporal to spatial domains, providing the genuine uncertainty information required for robust scientific inference.
\end{abstract}

\section{Introduction}

Inverse problems in scientific computing require extracting underlying physical parameters from experimental observations, a fundamental challenge across domains from medical imaging to materials characterization~\cite{stuart2010inverse,tarantola2005inverse}.
These problems are characterized by inherent parameter ambiguities where multiple parameter combinations can produce nearly identical observations, necessitating probabilistic approaches that quantify parameter uncertainty rather than providing false confidence through point estimates~\cite{kaipio2005statistical,stuart2010inverse}.

Various uncertainty quantification methods have been developed for inverse problems, including Bayesian neural networks~\cite{blundell2015weight}, Monte Carlo dropout\cite{gal2016dropout}, deep ensembles~\cite{lakshminarayanan2017simple}, and normalizing flows~\cite{ardizzone2018analyzing}.
Recently, the Conditional Diffusion Model-based Inverse Problem Solver (CDI) framework has been proposed for probabilistic inverse problem solving~\cite{cdiframework}.

CDI employs diffusion models~\cite{ho2020denoising} to generate parameter distributions conditioned on observational data, providing natural uncertainty quantification through posterior sampling.
Diffusion models have demonstrated strong empirical performance across generative modeling tasks~\cite{dhariwal2021diffusion,ramesh2022hierarchical} with relatively straightforward training procedures.
However, CDI has been demonstrated only for one-dimensional temporal signal processing in power system analysis, leaving open whether the framework generalizes to other data modalities—particularly spatial inverse problems prevalent across scientific domains.

\begin{figure*}[htb]
    \centering
    \begin{subfigure}[b]{0.20\textwidth}
        \centering
        \includegraphics[width=\textwidth]{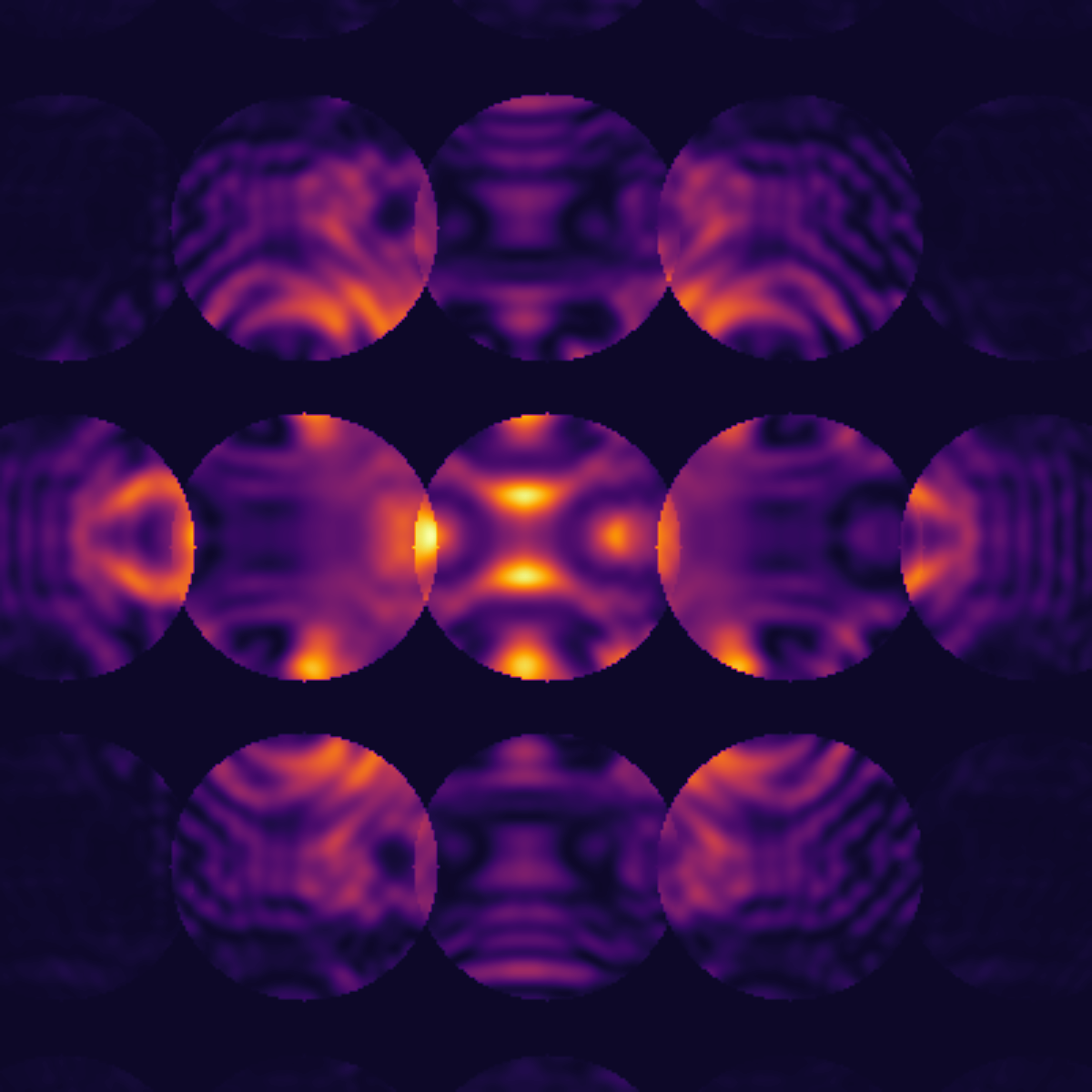}
    \end{subfigure}
    \hspace{0.01\textwidth}
    \begin{subfigure}[b]{0.20\textwidth}
        \centering
        \includegraphics[width=\textwidth]{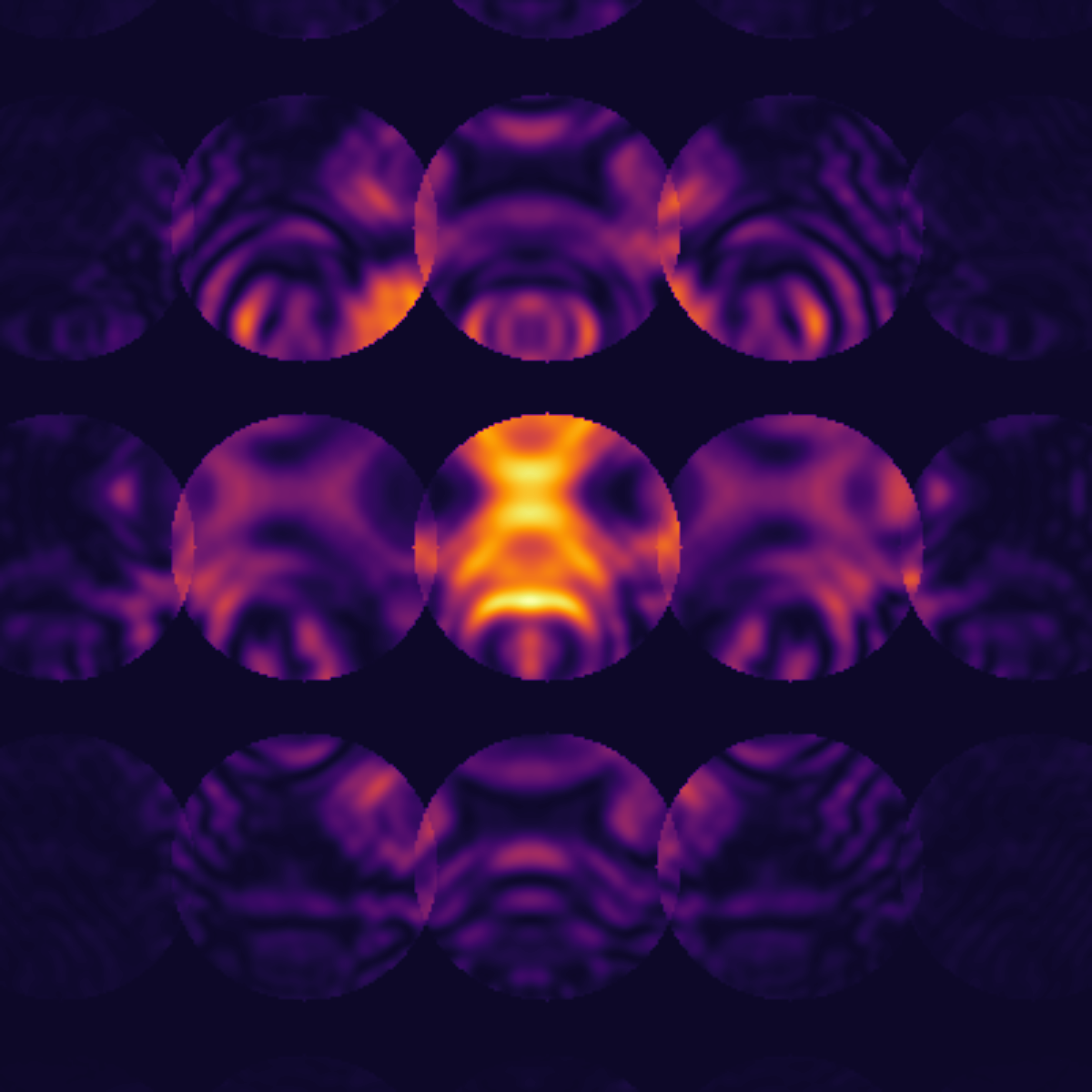}
    \end{subfigure}
    \hspace{0.01\textwidth}
    \begin{subfigure}[b]{0.20\textwidth}
        \centering
        \includegraphics[width=\textwidth]{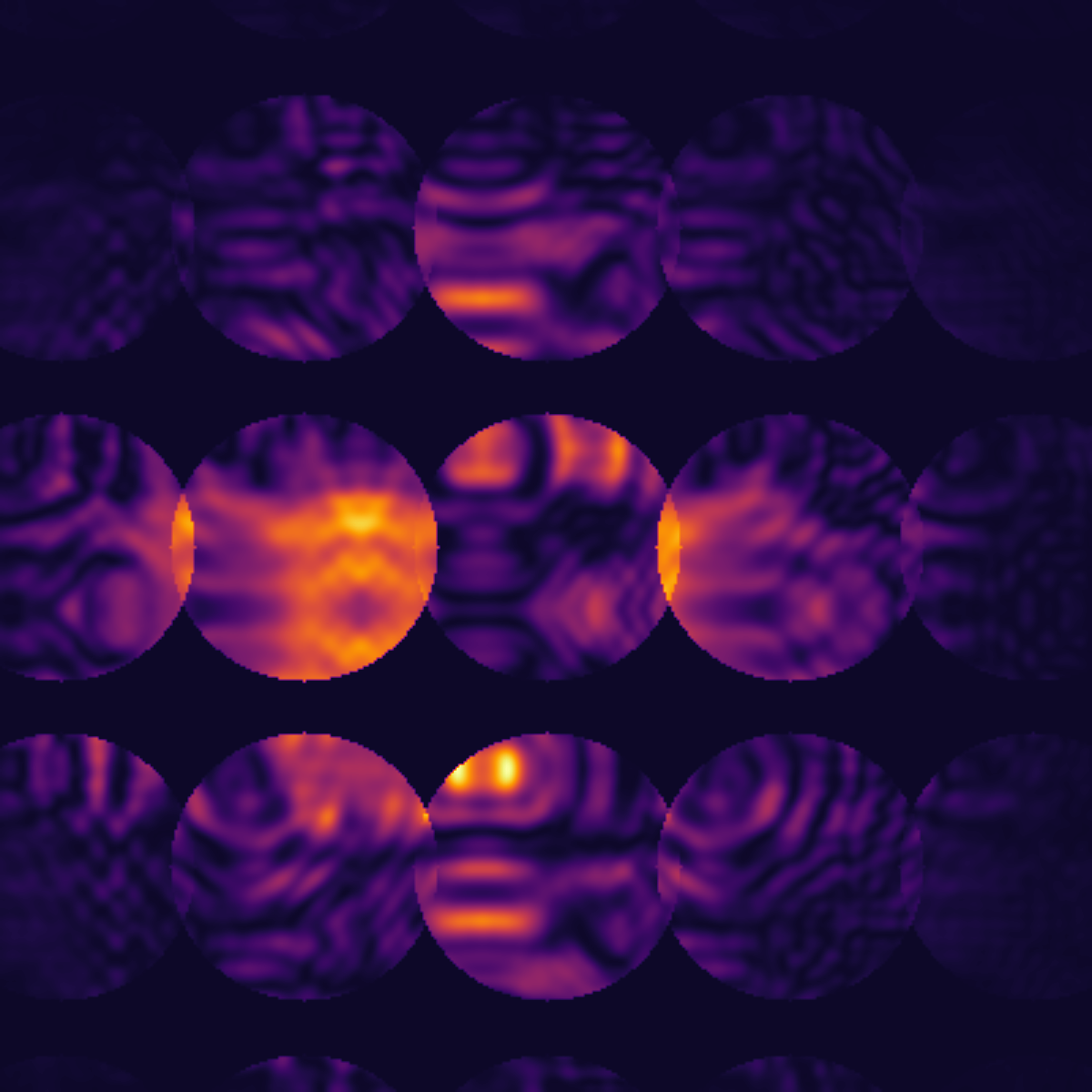}
    \end{subfigure}
    \hspace{0.01\textwidth}
    \begin{subfigure}[b]{0.20\textwidth}
        \centering
        \includegraphics[width=\textwidth]{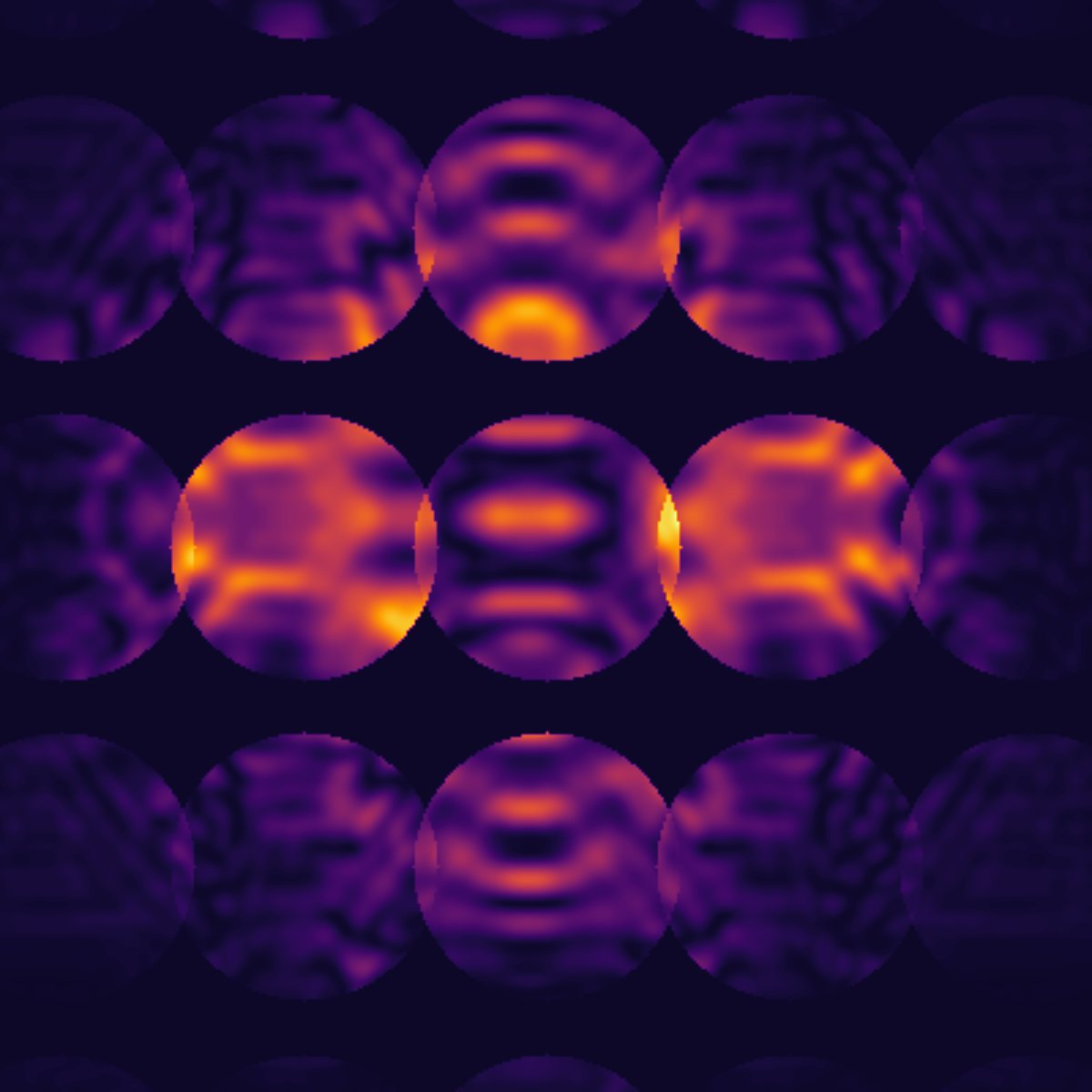}
    \end{subfigure}
    \caption{Representative CBED patterns from the simulated dataset. The patterns exhibit intensity variations that encode crystal thickness, tilt angles, Debye-Waller factors, and structure factors.
    }
    \label{fig:cbed_examples}
\end{figure*}

In this work, we investigate whether CDI generalizes to two-dimensional spatial conditioning.
We use convergent beam electron diffraction (CBED) parameter inference as our test domain -- a multi-parameter inverse problem in materials characterization where sample thickness, crystal orientation, thermal vibration coefficients, and electronic structure factors must be extracted from 2D diffraction patterns (Figure~\ref{fig:cbed_examples}).

The CBED problem exhibits characteristics valuable for uncertainty quantification evaluation: the significantly varying physical sensitivity of diffraction patterns to different parameters creates a natural spectrum from well-constrained quantities (sample geometry, structure factors) to poorly-constrained parameters (certain thermal coefficients), enabling systematic assessment of how uncertainty estimates reflect genuine measurement constraints.

Our contributions include:
\begin{itemize}
  \item We demonstrate successful extension of CDI from 1D temporal to 2D spatial observations, establishing framework applicability across data modalities
  \item We develop cross-modal transformer architecture enabling parameter-space diffusion conditioned on image observations
  \item We show that regression methods mask genuine parameter uncertainty by defaulting to training distribution means for poorly-constrained quantities, while CDI posteriors correctly reflect measurement constraints
\end{itemize}

These results establish that CDI provides principled uncertainty quantification for spatial inverse problems, successfully generalizing beyond its original temporal signal domain.
By validating on CBED parameter inference -- where varying parameter sensitivities create systematic identifiability challenges -- we demonstrate that diffusion-based approaches correctly distinguish well-constrained quantities from ambiguous parameters, providing measurement-specific uncertainty information that point estimate methods systematically mask.
This capability is essential for scientific applications where understanding parameter reliability is as critical as the estimates themselves.

\section{Related Work}

As discussed in the introduction, various uncertainty quantification methods have been developed for inverse problems, each with distinct trade-offs. Bayesian neural networks provide principled uncertainty through weight distributions but require specialized training~\cite{blundell2015weight}. Monte Carlo dropout offers simpler approximations with weaker guarantees~\cite{gal2016dropout}. Deep ensembles provide practical UQ through multiple models~\cite{lakshminarayanan2017simple}. Normalizing flows enable exact posteriors but impose architectural constraints~\cite{ardizzone2018analyzing}. Diffusion models have recently emerged as an alternative approach for uncertainty quantification in inverse problems, offering posterior sampling with relatively straightforward training procedures.

\subsection{Diffusion Models for Inverse Problems}

Diffusion models have recently been applied to inverse problems through posterior sampling approaches~\cite{songsolving,chung2022diffusion}.
These methods leverage pretrained diffusion models to sample from posterior distributions conditioned on observations, enabling uncertainty quantification without retraining.
Applications include image restoration, compressed sensing, and medical image reconstruction~\cite{kawar2022denoising,chung2022score,songsolving}.

The Conditional Diffusion Model-based Inverse Problem Solver (CDI) framework~\cite{cdiframework} takes a different approach: rather than conditioning on image observations to reconstruct images, CDI performs parameter-space diffusion conditioned on observational data to infer system parameters. This formulation naturally handles multi-parameter inverse problems where the goal is extracting physical quantities rather than reconstructing high-dimensional observations. CDI has demonstrated effective uncertainty quantification for one-dimensional temporal signal processing in power system analysis, but its applicability to higher-dimensional spatial data remained unexplored prior to this work.

\subsection{CBED Parameter Inference}

Convergent beam electron diffraction (CBED) analysis traditionally relies on iterative refinement methods that adjust parameters until simulated patterns match experimental observations~\cite{tsuda1999refinement,zuo1999direct}. Recent deep learning approaches have demonstrated successful parameter estimation through direct regression~\cite{xu2018deep,zhang2020atomic,yuan2021training}, achieving significant speedups over iterative methods. However, existing approaches provide only point estimates without uncertainty quantification, limiting their utility for scientific applications where parameter reliability assessment is essential. Our work addresses this gap by validating CDI's extension to spatial data on the CBED parameter inference problem.

\section{Methods}

\subsection{Problem Formulation}

Convergent beam electron diffraction (CBED) is an electron microscopy technique where a focused electron beam interacts with a crystalline sample, producing characteristic diffraction patterns that encode structural and electronic information. The resulting patterns consist of overlapping diffraction discs containing intensity modulations sensitive to sample thickness, crystal orientation, thermal vibrations, and valence electron distributions.

In this work, we consider CBED analysis of magnesium diboride (\chemfig{Mg B_2}) crystals, a well-characterized superconductor with hexagonal crystal structure. The analysis presents a multi-parameter inverse problem where the goal is to infer crystal and experimental parameters from observed 2D diffraction patterns.
The target parameters $\boldsymbol{\theta} = [\theta_1, \theta_2, ..., \theta_{13}]$ include six Debye-Waller factors ($u_{11}$, $u_{33}$, $u_{12}$ for both \chemfig{Mg} and \chemfig{B} atoms) encoding thermal atomic vibrations, two beam tilt angles ($\theta_x, \theta_y$) and sample thickness describing experimental geometry, and four low-order structure factors ($F_{001}$, $F_{100}$, $F_{101}$, $F_{002}$) encoding valence electron distributions.

CBED diffraction patterns exhibit significantly varying sensitivity across these parameters: geometric quantities (thickness, tilt angles) and structure factors produce strong, distinctive signatures in the diffraction intensity distributions, while Debye-Waller factors manifest as subtle intensity dampening that provides weak observational constraints. This parameter-dependent observational constraint spectrum makes CBED well-suited for evaluating whether uncertainty quantification methods correctly reflect measurement information content.

We generate training and evaluation datasets using physics-based Bloch wave simulations. Detailed simulation procedures, parameter ranges, and crystallographic considerations are provided in Appendix~\ref{sec:app_simulation}.

\begin{figure*}[htb]
    \centering
    \includegraphics[width=0.9\textwidth]{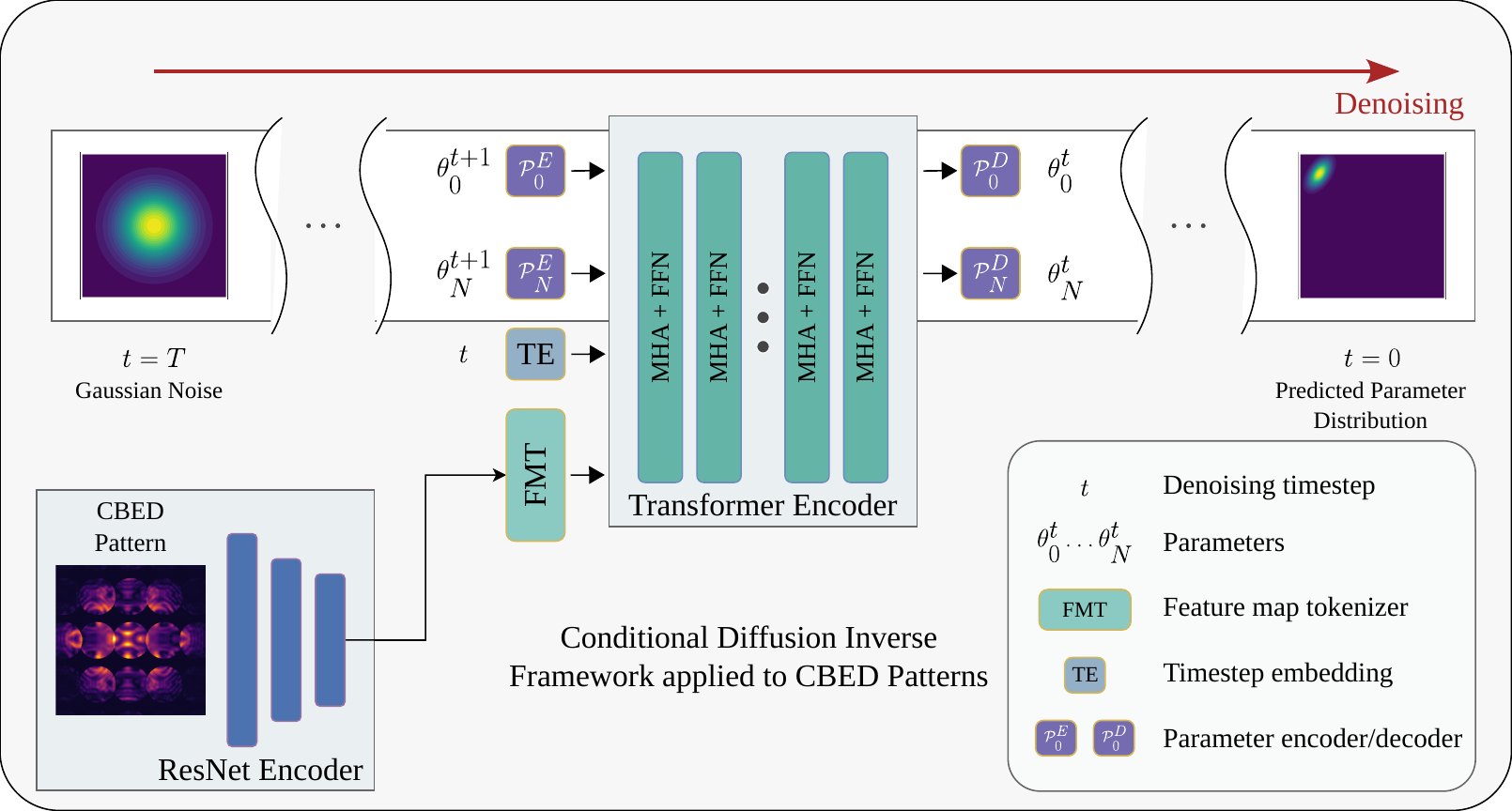}
    \caption{CDI architecture for CBED parameter inference.
    A ResNet-based encoder processes the input CBED pattern and extracts spatial features, which are tokenized by the feature map tokenizer (FMT).
    During reverse diffusion, the transformer encoder iteratively denoises parameter tokens from random noise ($t=T$) to clean estimates ($t=0$). 
    At each timestep, the transformer processes noisy parameter tokens conditioned on the image tokens and timestep embedding.
    Parameter encoder and decoder ($\mathcal{P}^E/\mathcal{P}^D$) map between parameter space and transformer embedding space.}
    \label{fig:cbed_diagram}
\end{figure*}

\subsection{CDI Framework for Inverse Problems}

The Conditional Diffusion Model-based Inverse Problem Solver (CDI) framework~\cite{cdiframework} demonstrates the ability to infer system parameter distributions and their joint correlations from observations of system behavior.
Unlike traditional supervised regression methods, which are only capable of providing single point parameter estimates, CDI leverages the inherent stochasticity of the diffusion model process and produces full posterior distributions of system parameters consistent with the data.
This enables parameter correlation studies and uncertainty quantification.

At its core, the CDI framework applies diffusion processes~\cite{sohl2015deep} directly in the parameter space ($\boldsymbol{\theta}$).
The framework employs a denoising diffusion probabilistic model~\cite{ho2020denoising} (DDPM), conditioned on observational data ($\mathbf{y}$), to generate parameter distributions consistent with the observations.
The DDPM consists of two processes: a forward diffusion process that gradually corrupts parameter vectors with Gaussian noise, and a learned reverse process that removes this noise to generate parameters consistent with observational data.

The forward diffusion process transforms the original parameter vector $\boldsymbol{\theta}_0$ into increasingly noisy versions $\boldsymbol{\theta}_1, \ldots, \boldsymbol{\theta}_T$ through a Markov chain:
\begin{equation}
q(\boldsymbol{\theta}_t|\boldsymbol{\theta}_{t-1}) = \mathcal{N}(\boldsymbol{\theta}_t; \sqrt{1-\beta_t} \boldsymbol{\theta}_{t-1}, \beta_t \mathbf{I})
\end{equation}
where $\{\beta_t\}_{t=1}^T$ is a variance schedule controlling the noise level at each step. Common choices include linear schedules $\beta_t = \beta_1 + (t-1)(\beta_T - \beta_1)/(T-1)$ or cosine schedules~\cite{nichol2021improved}. We employ a linear variance schedule for its demonstrated stability in parameter space diffusion.

The Gaussian nature of the forward process allows the Markov chain to be expressed in closed form:
\begin{equation}
q(\boldsymbol{\theta}_t|\boldsymbol{\theta}_0) = \mathcal{N}(\boldsymbol{\theta}_t; \sqrt{\bar{\alpha}_t} \boldsymbol{\theta}_0, (1-\bar{\alpha}_t)\mathbf{I})
\end{equation}
with $\alpha_t = 1 - \beta_t$ and $\bar{\alpha}_t = \prod_{s=1}^t \alpha_s$.

The reverse process learns to denoise parameters by predicting a less noisy version $\boldsymbol{\theta}_{t-1}$ from $\boldsymbol{\theta}_t$, conditioned on system observations $\mathbf{y}$:
\begin{equation}
p_\phi(\boldsymbol{\theta}_{t-1}|\boldsymbol{\theta}_t, \mathbf{y}) = \mathcal{N}(\boldsymbol{\theta}_{t-1}; \boldsymbol{\mu}_\phi(\boldsymbol{\theta}_t, \mathbf{y}, t), \boldsymbol{\Sigma}_\phi(\boldsymbol{\theta}_t, \mathbf{y}, t))
\end{equation}
In the DDPM model, the reverse process is parameterized by a neural network $\boldsymbol{\epsilon}_\phi(\boldsymbol{\theta}_t, \mathbf{y}, t)$ that predicts the noise added at step $t$ from the current noisy parameters $\boldsymbol{\theta}_t$ and system observations $\mathbf{y}$.
The mean $\boldsymbol{\mu}_\phi$ is then computed from the predicted noise and current noisy parameters $\boldsymbol{\theta}_t$, while the variance $\boldsymbol{\Sigma}_\phi$ follows a fixed schedule.

The neural network $\boldsymbol{\epsilon}_\phi(\boldsymbol{\theta}_t, \mathbf{y}, t)$ is trained to predict the noise $\boldsymbol{\epsilon}$ added during the forward process by minimizing the loss function below:
\begin{equation}
\mathcal{L} = \mathbb{E}_{t,\boldsymbol{\theta}_0,\boldsymbol{\epsilon}}[\|\boldsymbol{\epsilon} - \boldsymbol{\epsilon}_\phi(\boldsymbol{\theta}_t, \mathbf{y}, t)\|^2]
\end{equation}
This conditioning on $\mathbf{y}$ ensures generated parameters are consistent with the observational data.

At inference time, the learned denoising network enables posterior sampling: given observation $\mathbf{y}$, we generate multiple parameter samples $\{\boldsymbol{\theta}_0^{(i)}\}_{i=1}^K$ through repeated application of the reverse process, producing an empirical approximation to the posterior distribution $p(\boldsymbol{\theta}|\mathbf{y})$.

\subsection{Cross-Modal Transformer Architecture for Image-Based Inverse Problems}

The original CDI framework processes one-dimensional temporal signals through a hybrid CNN-transformer architecture. Extending this approach to two-dimensional spatial observations requires architectural innovations to handle the substantially higher dimensionality and spatial structure of image data while maintaining the parameter-space diffusion mechanism.

We employ a modified ResNet-34 backbone~\cite{he2016deep} as our vision encoder, processing CBED patterns of shape $(1, 320, 320)$ to extract hierarchical spatial features. We extend the standard architecture with an additional downsampling layer followed by a residual block to achieve a final spatial stride of 64, producing feature maps of shape $(512, 5, 5)$ (channels $\times$ height $\times$ width), capturing both local diffraction disc structures and global pattern geometry essential for parameter inference. These spatial feature maps are flattened along spatial dimensions to create sequences of length $L = H \times W = 25$, then linearly projected through $\mathbf{W}^{\text{vision}} \in \mathbb{R}^{512 \times d}$ to generate vision tokens $\mathbf{t}^{\text{vision}} \in \mathbb{R}^{25 \times d}$.

A key architectural contribution is the unified token representation enabling joint processing of heterogeneous data types. Each physical parameter $\theta_i$ is individually tokenized through learnable projections $\mathbf{W}^{\text{param}}_i \in \mathbb{R}^{1 \times d}$, producing parameter tokens $\mathbf{t}^{\text{param}}_i \in \mathbb{R}^{d}$. This per-parameter tokenization enables the attention mechanism to learn parameter-specific dependencies on spatial image features. Diffusion timestep $t$ is encoded through sinusoidal embeddings~\cite{vaswani2017attention} into temporal tokens $\mathbf{t}^{\text{time}} \in \mathbb{R}^{d}$, providing diffusion-stage-dependent conditioning.

The concatenated sequence $[\mathbf{t}^{\text{params}}, \mathbf{t}^{\text{vision}}, \mathbf{t}^{\text{time}}]$ is processed through $N=6$ transformer encoder layers with dimension $d=512$ and 8 attention heads, following standard transformer configurations~\cite{vaswani2017attention}. This architecture enables bidirectional cross-modal attention: parameter tokens attend to relevant spatial image features while vision tokens provide context-dependent conditioning for parameter denoising. Unlike standard image regression architectures that process images through feedforward pathways, our approach allows each parameter being inferred to dynamically weight its dependence on spatially-localized diffraction features through learned attention patterns. The transformer outputs denoised parameter token representations $\tilde{\mathbf{t}}^{\text{param}}_i$, which are decoded to parameter space through learned linear projections, producing the noise prediction $\boldsymbol{\epsilon}_\phi(\boldsymbol{\theta}_t, \mathbf{I}, t)$ for the diffusion denoising objective.

\subsection{Uncertainty Quantification Evaluation}

The CDI framework generates parameter distributions conditioned on observational data. However, the quality of these probabilistic outputs must be rigorously assessed to confirm that the resulting uncertainty estimates are meaningful.
The uncertainty quantification literature~\cite{kuleshov2018accurate,gneiting2007probabilistic} provides established metrics for evaluating distributional predictions, which we employ to validate our approach.

\textbf{Calibration.} Calibration measures the alignment between predicted confidence and empirical accuracy.
For a well-calibrated model, when the model predicts $90\%$ confidence that parameters should be contained within a given region, the true parameter should fall within that interval $90\%$ of the time.
We assess calibration through reliability diagrams that plot observed frequencies of true parameters falling within prediction intervals against the predicted confidence levels.

Formally, for a given observation $\mathbf{y}_i$ and confidence level $\alpha$, one can construct a confidence region $\text{CI}_\alpha(\mathbf{y}_i)$ that should contain the true parameter value $\hat\theta_i$ with probability $\alpha$ according to the model predictions.
The empirical coverage probability measures the actual frequency at which true parameters fall within their predicted regions:
\begin{equation}
\text{Coverage}_\alpha =
    \frac{1}{N}\sum_{i=1}^N
        \mathbb{I}\left[\hat\theta_i \in \text{CI}_\alpha(\mathbf{y}_i)\right]
\end{equation}
where $N$ is the size of the dataset and $\mathbb{I}$ is an indicator function. For a perfectly calibrated model, the coverage probability should equal the nominal confidence level $\alpha$. Deviations indicate miscalibration.

\textbf{Sharpness.} Sharpness quantifies the precision of uncertainty estimates by measuring the width of prediction intervals. Sharp predictions provide narrow confidence intervals while maintaining proper coverage. We calculate sharpness as the average interval width:
\begin{equation}
\text{Sharpness}_\alpha = \frac{1}{N}\sum_{i=1}^N \left(
    \text{CI}^\text{upper}_\alpha(\mathbf{y}_i)
    -
    \text{CI}^\text{lower}_\alpha(\mathbf{y}_i)
\right)
\end{equation}
where $\text{CI}^\text{upper}_\alpha(\mathbf{y}_i)$ and $\text{CI}^\text{lower}_\alpha(\mathbf{y}_i)$ are the upper and lower bounds of the confidence interval for observation $\mathbf{y}_i$.

These metrics collectively distinguish between models that provide genuinely informative uncertainty estimates versus those with systematic biases or no uncertainty quantification. Regression approaches produce point estimates without uncertainty intervals, failing to capture parameter-dependent measurement precision inherent in inverse problems.

\section{Experimental Setup}

\textbf{Dataset.} Rigorous uncertainty quantification evaluation requires verified ground truth parameter labels to assess calibration quality and coverage probability. We employ physics-based Bloch wave simulations that generate CBED patterns with known parameter values, enabling systematic quantitative assessment of probabilistic predictions. We generate 1M training patterns and 100K test patterns by uniformly sampling parameters from physically realistic ranges detailed in Appendix~\ref{sec:app_simulation}.

\textbf{Baselines.} We compare CDI against direct regression baselines to establish performance context. We evaluate multiple CNN architectures (ResNet~\cite{he2016deep}, EfficientNet~\cite{tan2019efficientnet}) developed for natural image tasks to ensure our findings are robust across different network designs rather than architecture-specific. All regression models are trained to minimize L2 loss between predicted and true parameter values. Training configurations are detailed in Appendix~\ref{sec:app_training}.

\textbf{Training.} All models use AdamW optimizer. Complete training configurations including learning rates, batch sizes, and learning rate schedules are provided in Appendix~\ref{sec:app_training}.

\section{Results}

We evaluate our approach on a test set of 100K simulated patterns with known ground truth parameters, enabling rigorous uncertainty quantification validation

\subsection{Method Performance}

We evaluate our \thename approach against direct regression baselines using normalized root mean square error (NRMSE) for cross-parameter comparison, where $\text{NRMSE}(\theta) = \text{RMSE}(\theta)/(\theta_\text{max} - \theta_\text{min})$ normalizes errors by parameter range for meaningful comparison across different physical quantities.
Notably, \thename's higher NRMSE on certain parameters reflects correct uncertainty quantification for poorly-constrained quantities rather than inferior accuracy -- a distinction analyzed in Section~\ref{sec:identifiability}.
Table~\ref{tab:nrmse_comparison} presents performance results across all 13 parameters.

\begin{table*}[htb]
    \centering
    \resizebox{\textwidth}{!}{%
    \begin{tabular}{|c|c|c|c|c|c|c|c|c|c|c|c|c|c|}
\hline
\multirow{2}{*}{\textbf{Model}}
    & \multicolumn{3}{c|}{\textbf{DW Factors (Mg)}}
    & \multicolumn{3}{c|}{\textbf{DW Factors (B)}} 
    & \multicolumn{4}{c|}{\textbf{Structure Factors}}
    & \multicolumn{2}{c|}{\textbf{Tilt Angles}}
    & \textbf{Sample} \\
\cline{2-13}
  & $u_{11}$ & $u_{33}$ & $u_{12}$
  & $u_{11}$ & $u_{33}$ & $u_{12}$
  & $F_{001}$ & $F_{100}$ & $F_{101}$ & $F_{002}$
  & $\theta_x$ & $\theta_y$ & \textbf{Thickness} \\
\hline
\rowcolor{gray!15}
ResNet-10 & 0.157 & 0.189 & 0.283 & 0.211 & 0.229 & 0.287 & 0.034 & 0.047 & 0.054 & 0.074 & 0.018 & 0.017 & 0.036 \\
\rowcolor{white}
ResNet-18 & 0.140 & 0.156 & 0.281 & 0.182 & 0.198 & 0.286 & 0.024 & 0.036 & 0.041 & 0.061 & 0.013 & 0.013 & 0.030 \\
\rowcolor{gray!15}
ResNet-34 & 0.127 & 0.138 & 0.279 & 0.168 & 0.182 & 0.285 & 0.020 & 0.031 & 0.038 & 0.054 & 0.011 & 0.011 & 0.034 \\
\rowcolor{white}
ResNet-50 & 0.129 & 0.139 & 0.277 & 0.165 & 0.176 & 0.285 & 0.020 & 0.030 & 0.037 & 0.053 & 0.010 & 0.010 & 0.031 \\
\hline
\rowcolor{gray!15}
EfficientNet-B0 & 0.130 & 0.166 & 0.281 & 0.181 & 0.212 & 0.286 & 0.023 & 0.037 & 0.042 & 0.065 & 0.013 & 0.015 & 0.027 \\
\rowcolor{white}
EfficientNet-B1 & 0.132 & 0.164 & 0.281 & 0.175 & 0.204 & 0.286 & 0.022 & 0.036 & 0.044 & 0.061 & 0.011 & 0.010 & 0.028 \\
\rowcolor{gray!15}
EfficientNet-B2 & 0.124 & 0.151 & 0.281 & 0.172 & 0.196 & 0.286 & 0.021 & 0.033 & 0.042 & 0.061 & 0.010 & 0.009 & 0.034 \\
\hline
\rowcolor{white}
CDI & 0.167 & 0.190 & 0.402 & 0.230 & 0.253 & 0.401 & 0.020 & 0.031 & 0.042 & 0.061 & 0.008 & 0.008 & 0.022 \\
\hline
    \end{tabular}
    }
    \caption{NRMSE Performance Comparison. Parameters are organized by physical quantity: Debye-Waller factors for Mg ($u_{11}, u_{33}, u_{12}$), Debye-Waller factors for B ($u_{11}, u_{33}, u_{12}$), structure factors ($F_{001}, F_{100}, F_{101}, F_{002}$), beam tilt angles $(\theta_x, \theta_y)$, and sample thickness. Lower NRMSE indicates better prediction accuracy.\\
    Note: For uniformly distributed parameters, theoretical NRMSE values are $1/\sqrt{12} \approx 0.289$ when predicting distribution mean (regression behavior on unconstrained parameters) and $1/\sqrt{6} \approx 0.408$ when sampling uniformly (diffusion behavior on unconstrained parameters). Values near these thresholds indicate poorly-constrained parameters. See Section~\ref{sec:identifiability} for detailed analysis.
    }
    \label{tab:nrmse_comparison}
\end{table*}

Regression architectures demonstrate the expected trend of improving performance with increased model capacity, consistent with scaling patterns observed in computer vision literature.
Notably, all regression methods achieve excellent accuracy on geometric parameters (thickness, tilt angles) and structure factors (NRMSE $< 0.07$) while showing systematically higher errors on Debye-Waller factors (NRMSE $0.13-0.28$).
\thename achieves the best performance on geometric parameters (thickness, tilt angles) and matches regression accuracy on structure factors, while showing higher NRMSE values on Debye-Waller factors. This reflects fundamental differences in handling poorly-constrained parameters, analyzed below.

To understand these error patterns, we examine how each method handles parameters with varying observational constraints. For this and subsequent analyses, we compare \thename against ResNet-50 as it achieves the best overall regression performance.

\subsection{Parameter Identifiability Analysis}
\label{sec:identifiability}

\thename's higher NRMSE values on Debye-Waller factors compared to regression reveal a fundamental difference in how these methods handle poorly-constrained parameters rather than indicating inferior performance.
Debye-Waller factors encode thermal atomic vibrations that manifest as subtle intensity dampening in diffraction patterns.
Under the experimental conditions simulated in this work, Debye-Waller factors produce weak observable signatures, making these parameters difficult to constrain from the diffraction data alone.

\begin{figure}[htbp]
    \centering
    \hfill
        \begin{subfigure}[b]{0.49\textwidth}
            \includegraphics[width=\textwidth]{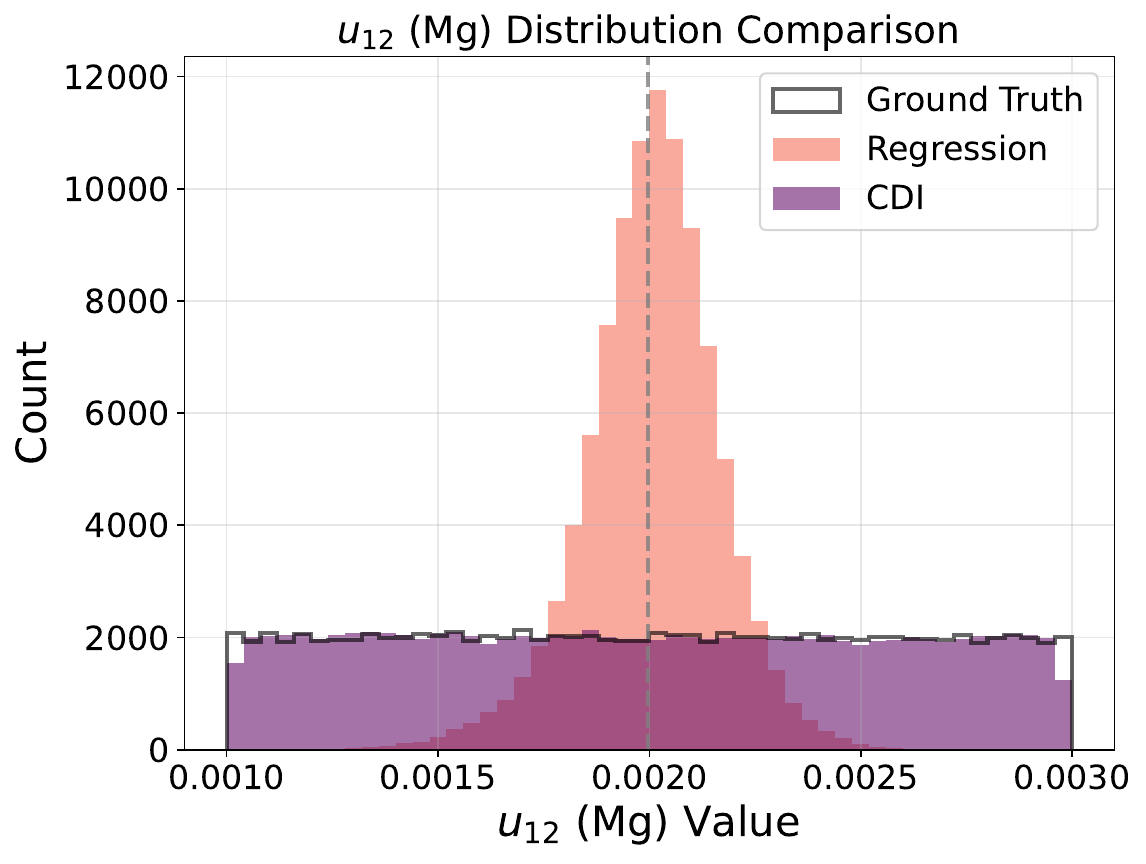}
        \end{subfigure}
        \begin{subfigure}[b]{0.48\textwidth}
            \includegraphics[width=\textwidth]{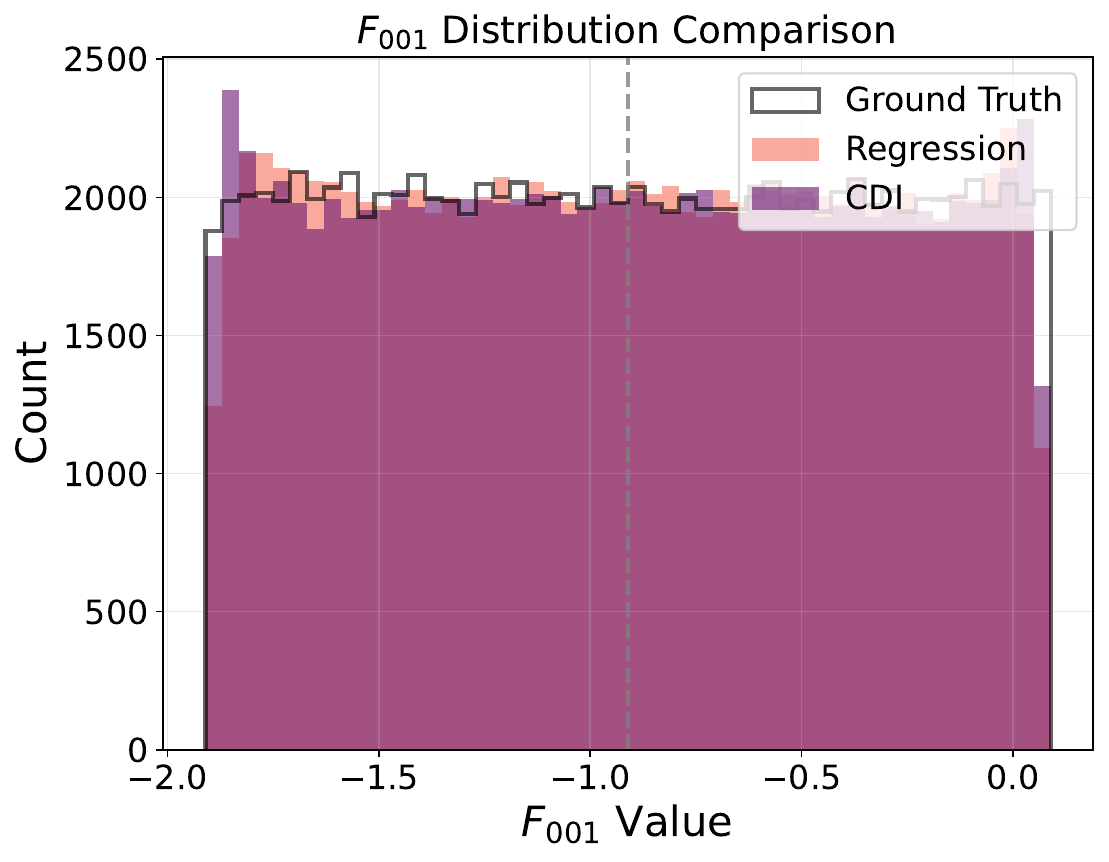}
        \end{subfigure}
    \hfill
    \caption{Parameter distribution comparison for unconstrained and well-constrained parameters. \textbf{Left:} Debye-Waller factor $u_{12}$ (Mg) distributions from ground truth (uniform), regression predictions, and \thename samples. \textbf{Right:} Structure factor $F_{001}$ distributions from the same three sources. Distributions are constructed from 100,000 test samples.}
    \label{fig:distribution_bias}
\end{figure}

In general, for parameters that cannot be inferred from the inputs, regression methods learn to predict the mean of the parameters, calculated from the training set. For a training dataset with a uniform parameter distribution, predicting its mean will result in a theoretical NRMSE of $1/\sqrt{12} \approx 0.289$.
Debye-Waller factors $u_{12}$ in Table~\ref{tab:nrmse_comparison} have values very close to $0.28$, suggesting that the regression model is simply outputting the distribution mean instead of performing any meaningful inference.

Similarly, for parameters that cannot be constrained by the inputs, diffusion models reproduce the training distribution of parameters rather than converging to a point estimate.
For a training dataset with a uniform parameter distribution, a diffusion model will generate parameters uniformly at random, resulting in NRMSE of $1/\sqrt{6} \approx 0.408$.
The observed NRMSE values for \thename on Debye-Waller factor $u_{12}$ of $0.40$ closely match this theoretical prediction, suggesting that \thename reflects the inherent parameter uncertainty when observational constraints are weak.

We can verify this analysis by examining the actual parameter distributions produced by regression and \thename methods (Figure~\ref{fig:distribution_bias}).
For unconstrained parameters, we expect regression to produce sharp distributions concentrated at the training set mean, while \thename should generate broad distributions matching the ground truth parameter range.
For well-constrained parameters, both methods should produce distributions that successfully capture the true parameter values.

Figure~\ref{fig:distribution_bias} confirms these predictions. For the unconstrained Debye-Waller factor $u_{12}$, regression produces a sharp distribution concentrated at the training set mean, while \thename generates a broad uniform distribution matching the ground truth.
In contrast, for the well-constrained structure factor $F_{001}$, both methods successfully recover distributions close to the ground truth, demonstrating that \thename maintains accuracy when observational information exists while correctly reflecting uncertainty when it does not.

\subsection{Uncertainty Calibration and Quality}

Having established that \thename produces appropriate uncertainty for unconstrained parameters, we now assess the quality of its probabilistic predictions for all parameters through rigorous calibration analysis.
Well-calibrated uncertainty estimates are essential for scientific decision-making—confidence intervals should contain true values at their claimed rates, distinguishing reliable predictions from arbitrary parameter variability.

\begin{figure}[htbp]
    \centering
    \hfill
    \begin{subfigure}[b]{0.45\textwidth}
        \centering
        \includegraphics[width=\textwidth]{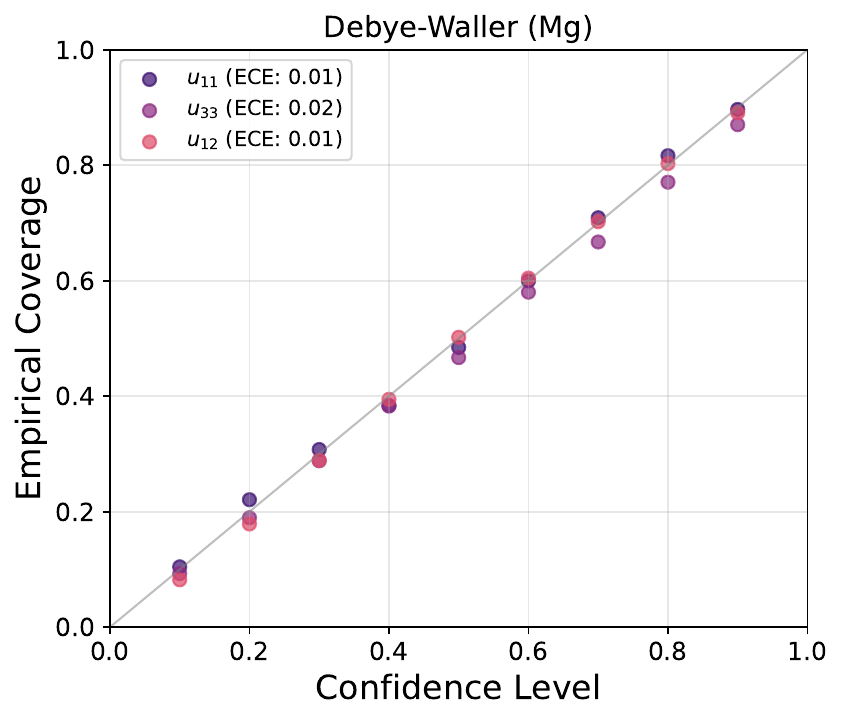}
    \end{subfigure}
    \hfill
    \begin{subfigure}[b]{0.45\textwidth}
        \centering
        \includegraphics[width=\textwidth]{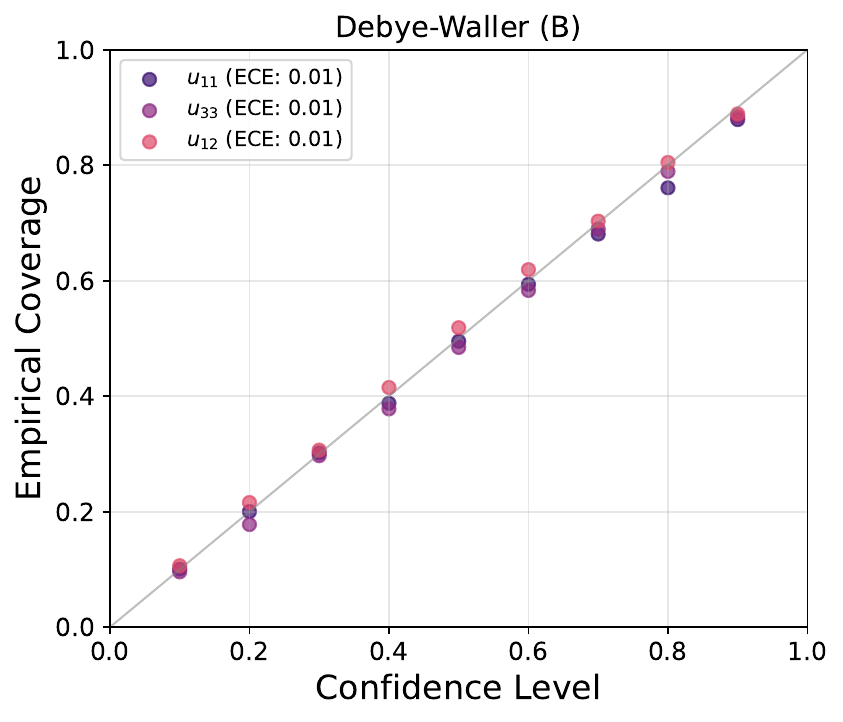}
    \end{subfigure}
    \hfill

    \vspace{1em} 

    \hfill
    \begin{subfigure}[b]{0.45\textwidth}
        \centering
        \includegraphics[width=\textwidth]{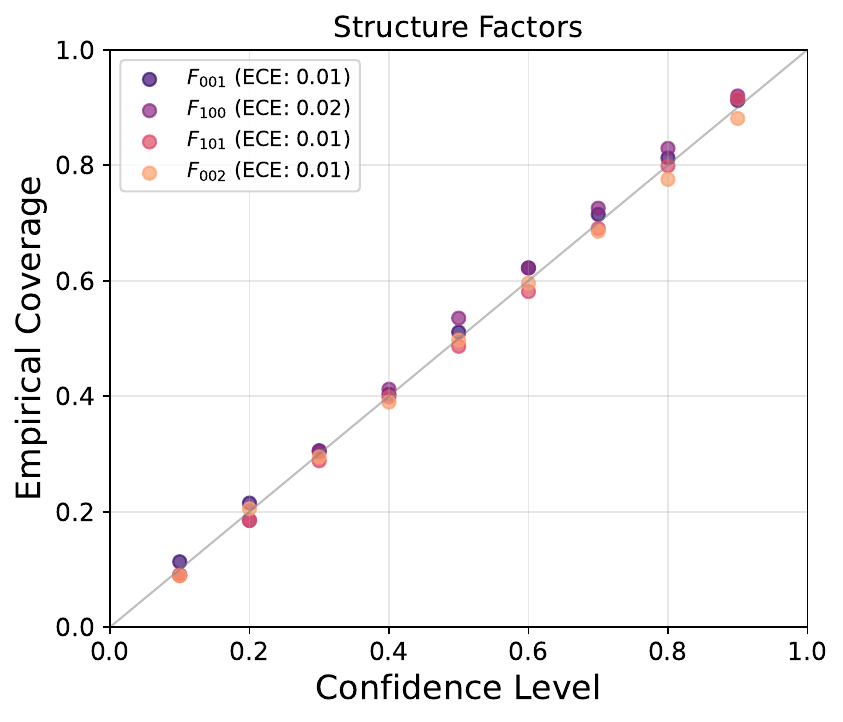}
    \end{subfigure}
    \hfill
    \begin{subfigure}[b]{0.45\textwidth}
        \centering
        \includegraphics[width=\textwidth]{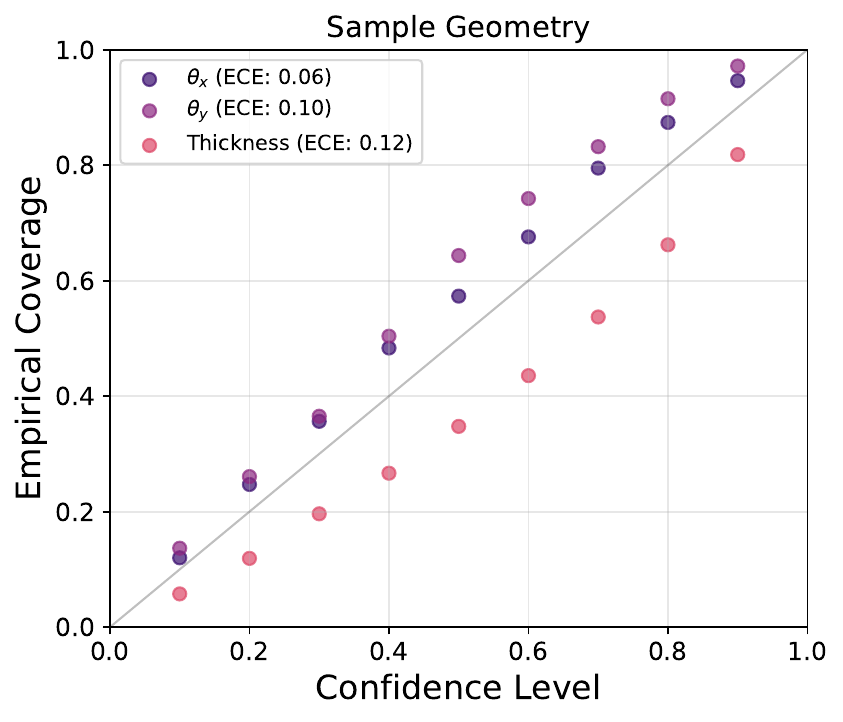}
    \end{subfigure}
    \hfill
    \caption{
    Calibration plots for CBED parameters showing empirical coverage frequencies vs predicted confidence levels. Diagonal lines indicate perfect calibration. (Top left) Debye-Waller Mg, (top right) Debye-Waller B, (bottom left) structure factors, (bottom right) sample geometry parameters.}
    \label{fig:uq_calibration}
\end{figure}

Figure~\ref{fig:uq_calibration} demonstrates calibration quality across CBED parameters.
The plots show empirical coverage frequencies versus predicted confidence levels.
Points on the diagonal indicate that confidence intervals contain true values at their claimed rates (e.g., $90\%$ intervals contain truth $90\%$ of the time); deviations indicate miscalibration.
Calibration errors remain within $\pm10\%$ of nominal levels across all parameters, meeting standard accuracy thresholds for probabilistic predictions.

\begin{figure}[htbp]
    \centering
    \hfill
    \begin{subfigure}[b]{0.45\textwidth}
        \centering
        \includegraphics[width=\textwidth]{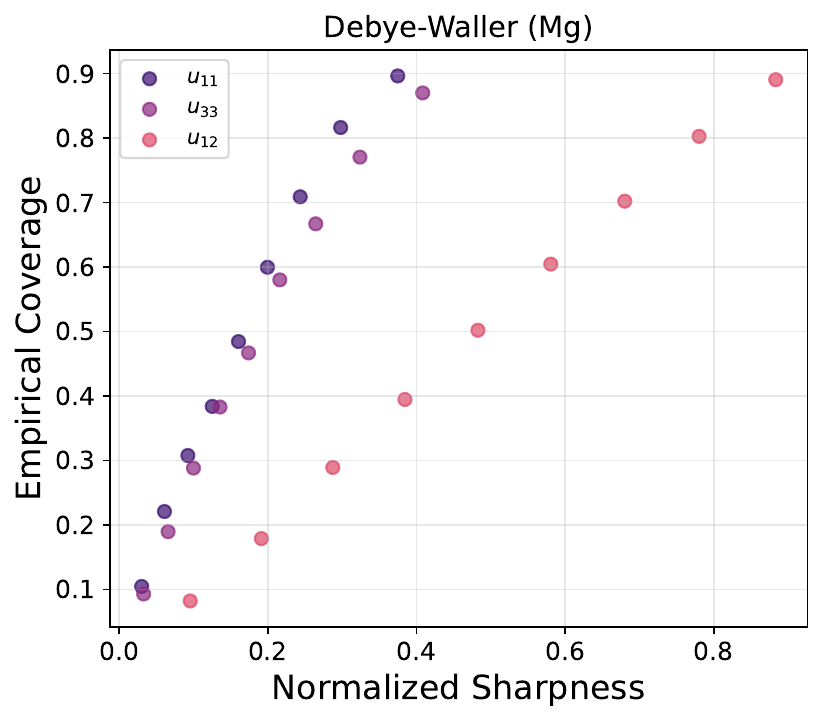}
    \end{subfigure}
    \hfill
    \begin{subfigure}[b]{0.45\textwidth}
        \centering
        \includegraphics[width=\textwidth]{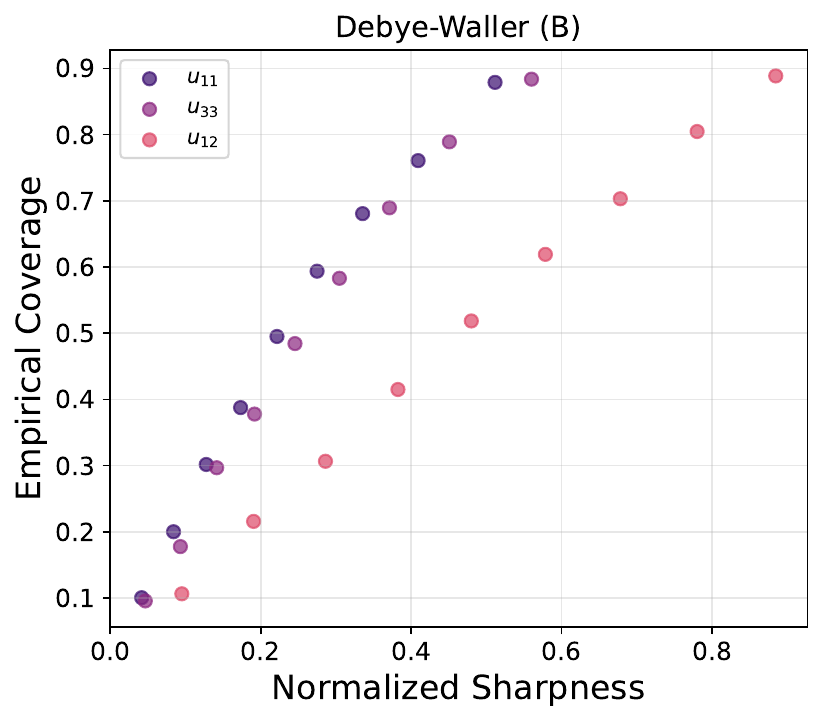}
    \end{subfigure}
    \hfill

    \vspace{1em} 

    \hfill
    \begin{subfigure}[b]{0.45\textwidth}
        \centering
        \includegraphics[width=\textwidth]{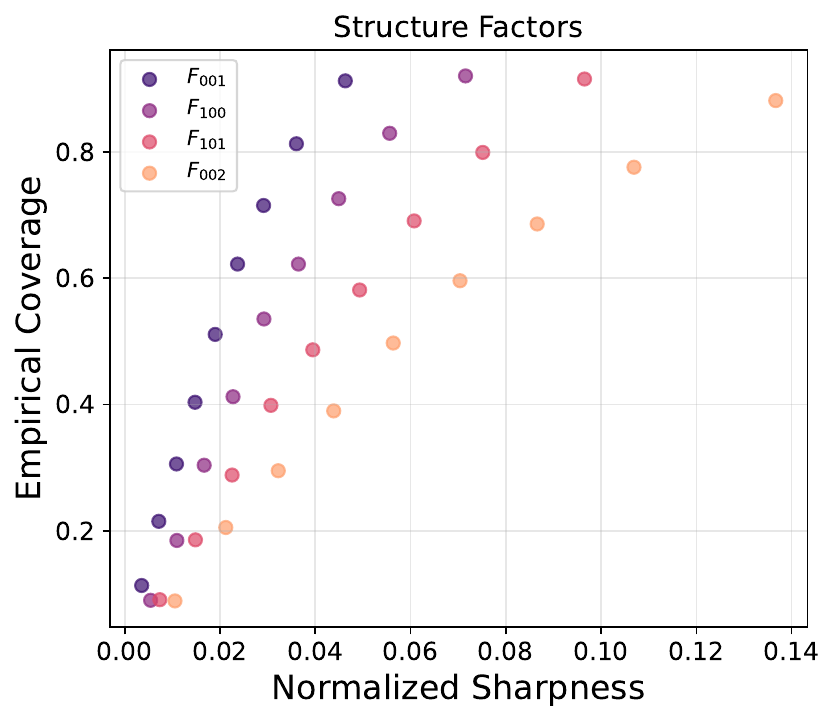}
    \end{subfigure}
    \hfill
    \begin{subfigure}[b]{0.45\textwidth}
        \centering
        \includegraphics[width=\textwidth]{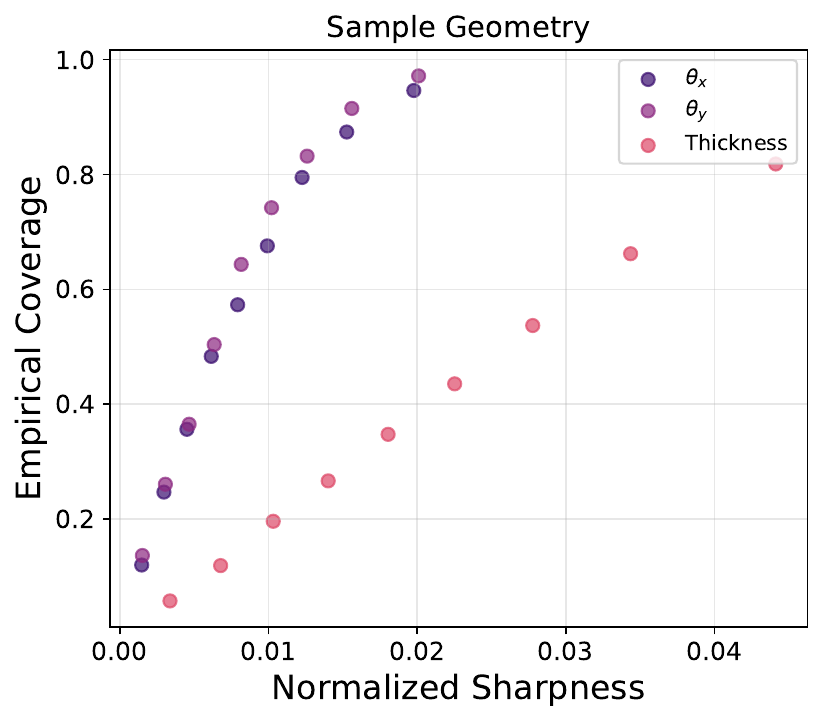}
    \end{subfigure}
    \hfill
    \caption{Coverage-sharpness trade-off analysis for representative CBED parameters. Each plot shows coverage probability (y-axis) versus prediction interval sharpness (x-axis) normalized over the parameter range at various confidence levels. (Top left) Debye-Waller Mg, (top right) Debye-Waller B, (bottom left) structure factors, (bottom right) sample geometry parameters.}
    \label{fig:uq_cov_sharp}
\end{figure}

Beyond calibration curves, the coverage-sharpness trade-off reveals prediction efficiency across parameter types (Figure~\ref{fig:uq_cov_sharp}). Normalized sharpness, measured as prediction interval width normalized by parameter range, quantifies how much of the parameter space must be captured to achieve target coverage levels.
Lower sharpness values (narrower intervals) indicate more informative predictions.

Sample geometry parameters demonstrate the strongest performance. Tilt angles achieve near-complete coverage with sharpness around $3\%$ of their parameter range, indicating highly accurate predictions.
Thickness shows slightly broader intervals but maintains favorable trade-offs.
Structure factors exhibit good prediction efficiency, with $10\%$ parameter range coverage yielding $50-100\%$ empirical coverage depending on the specific factor.
In contrast, Debye-Waller factors require substantially broader intervals to achieve equivalent coverage, with $u_{12}$ parameters exhibiting nearly 1:1 relationships between normalized sharpness and coverage.
This indicates less informative predictions where large fractions of the parameter space must be considered to capture true values reliably.

These results suggest that observational information content varies significantly across parameter types. Tilt angles and structure factors are well-constrained by CBED patterns, enabling sharp probabilistic predictions. Debye-Waller factors appear less directly constrained by the available diffraction information, resulting in broader uncertainty distributions that correctly reflect this ambiguity rather than producing overconfident narrow predictions.

\subsection{Qualitative Analysis: Regression vs CDI}

In the previous section, we demonstrated that CDI produces well-calibrated uncertainty estimates that correctly reflect parameter identifiability.  We now examine the characteristics of these distributions through concrete examples.

\begin{figure}[htbp]
\centering
\begin{minipage}[c]{0.30\textwidth}
    \centering
    \includegraphics[width=0.8\textwidth]{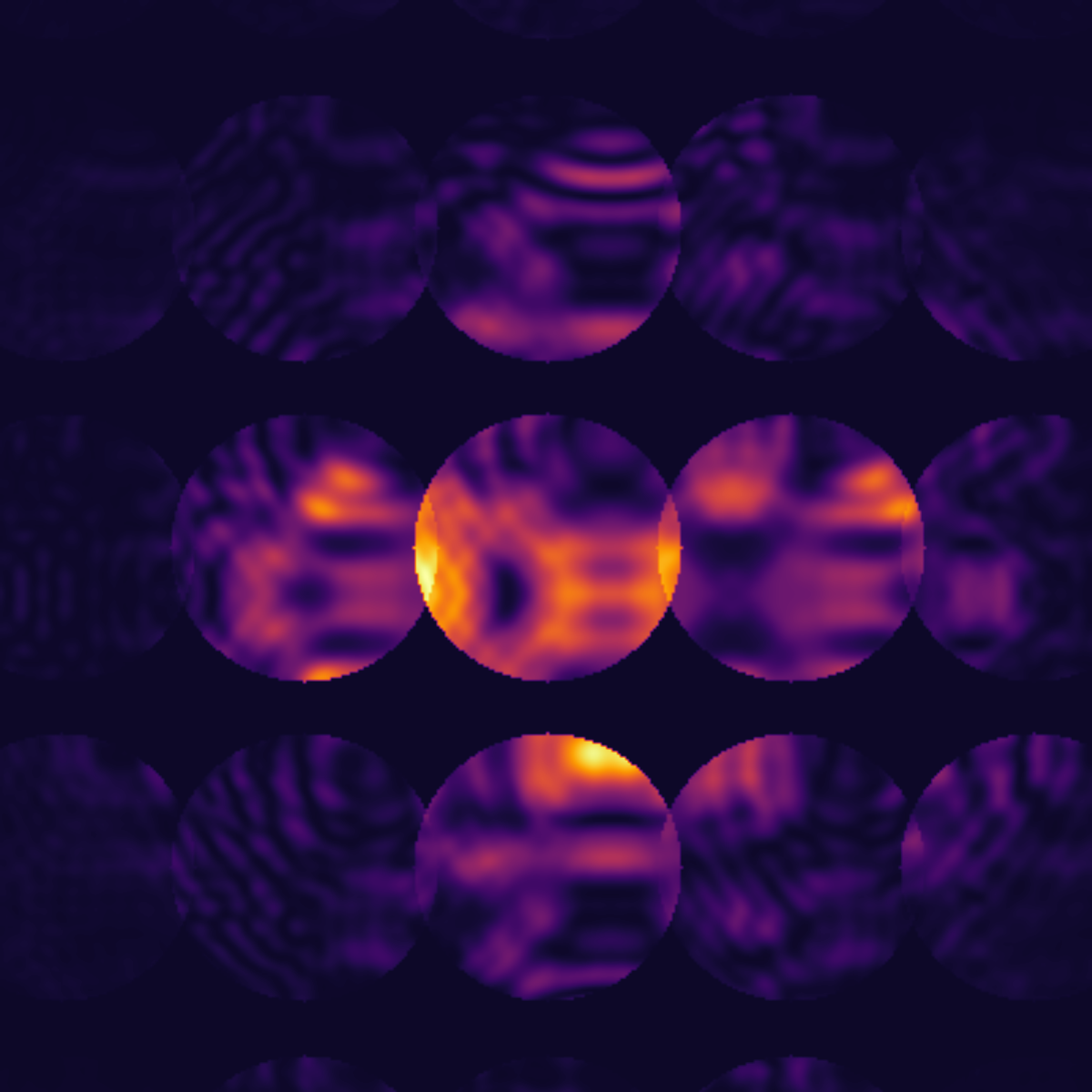}
    \caption*{(a) Input CBED Pattern}
\end{minipage}%
\begin{minipage}[c]{0.60\textwidth}
    \centering
    \begin{tabular}{ccc}
        \includegraphics[width=0.33\textwidth]{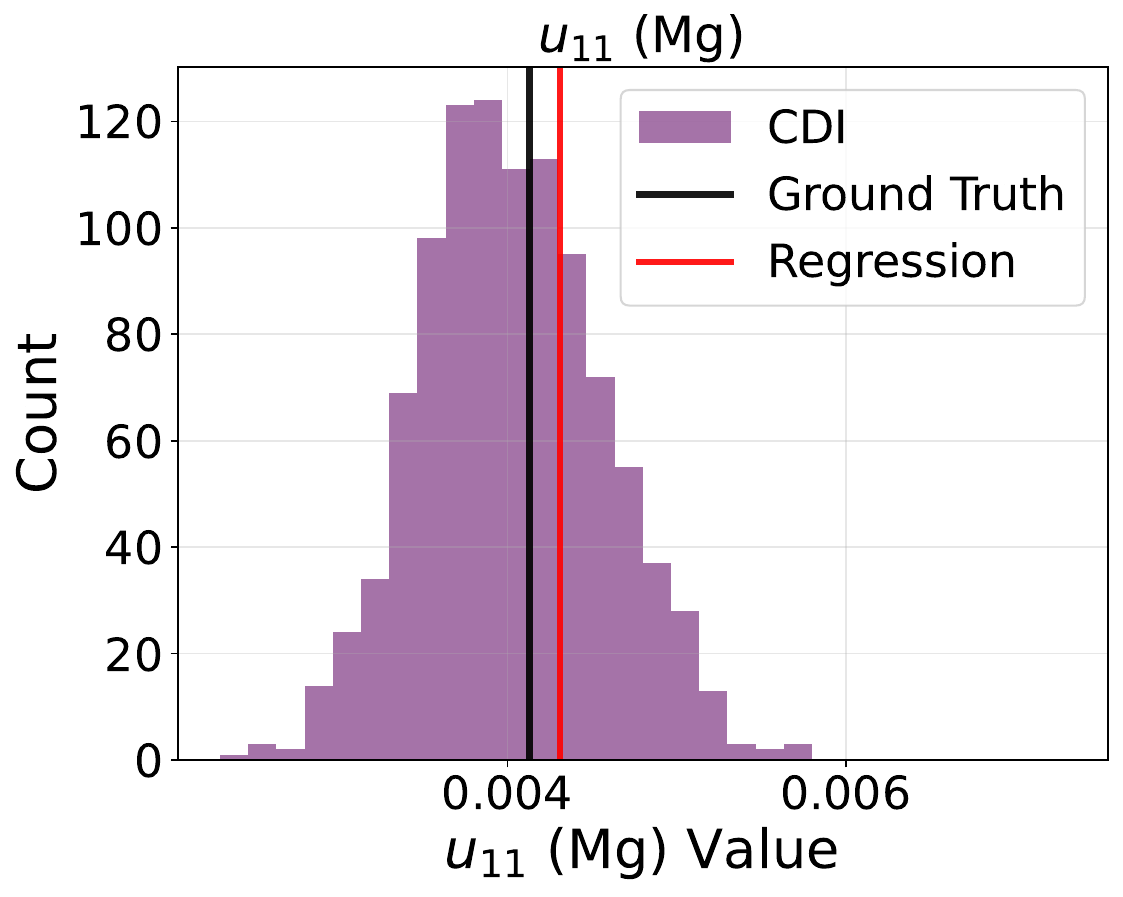} &
        \includegraphics[width=0.33\textwidth]{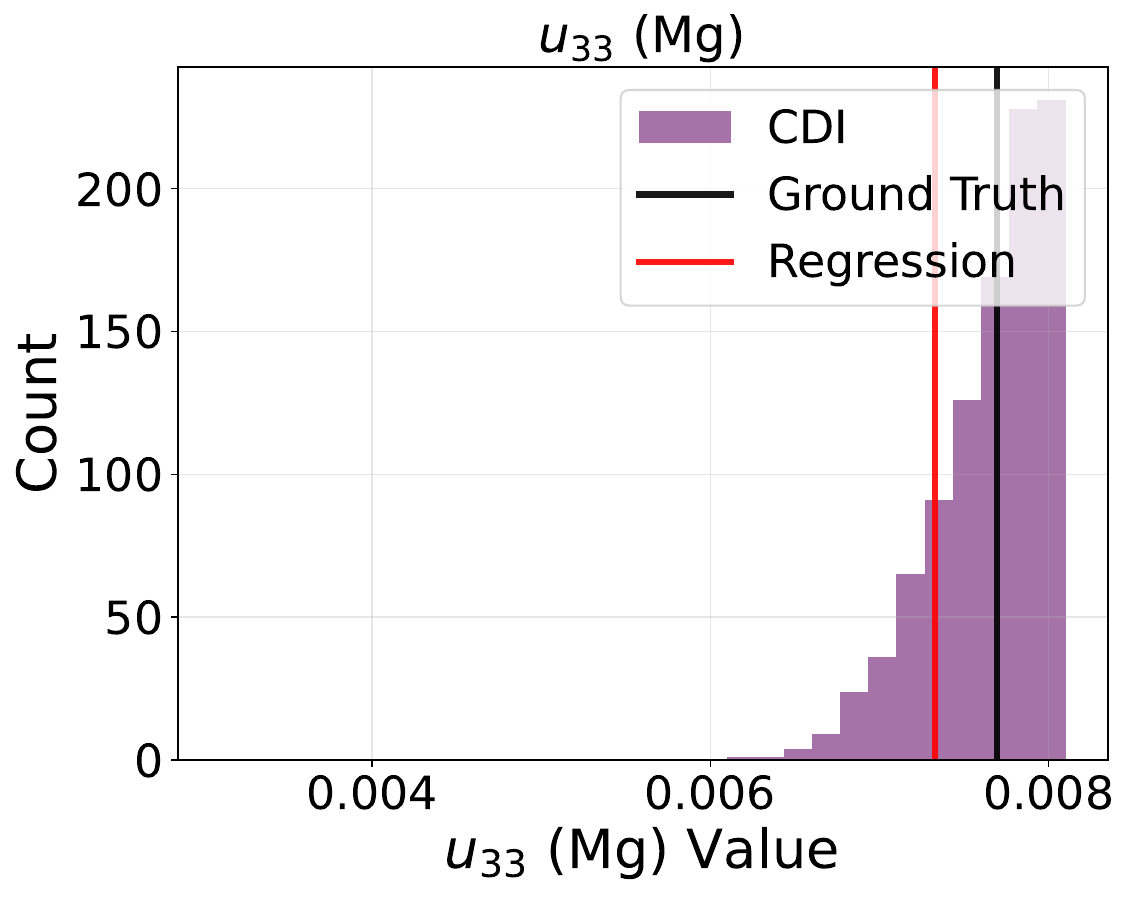} &
        \includegraphics[width=0.33\textwidth]{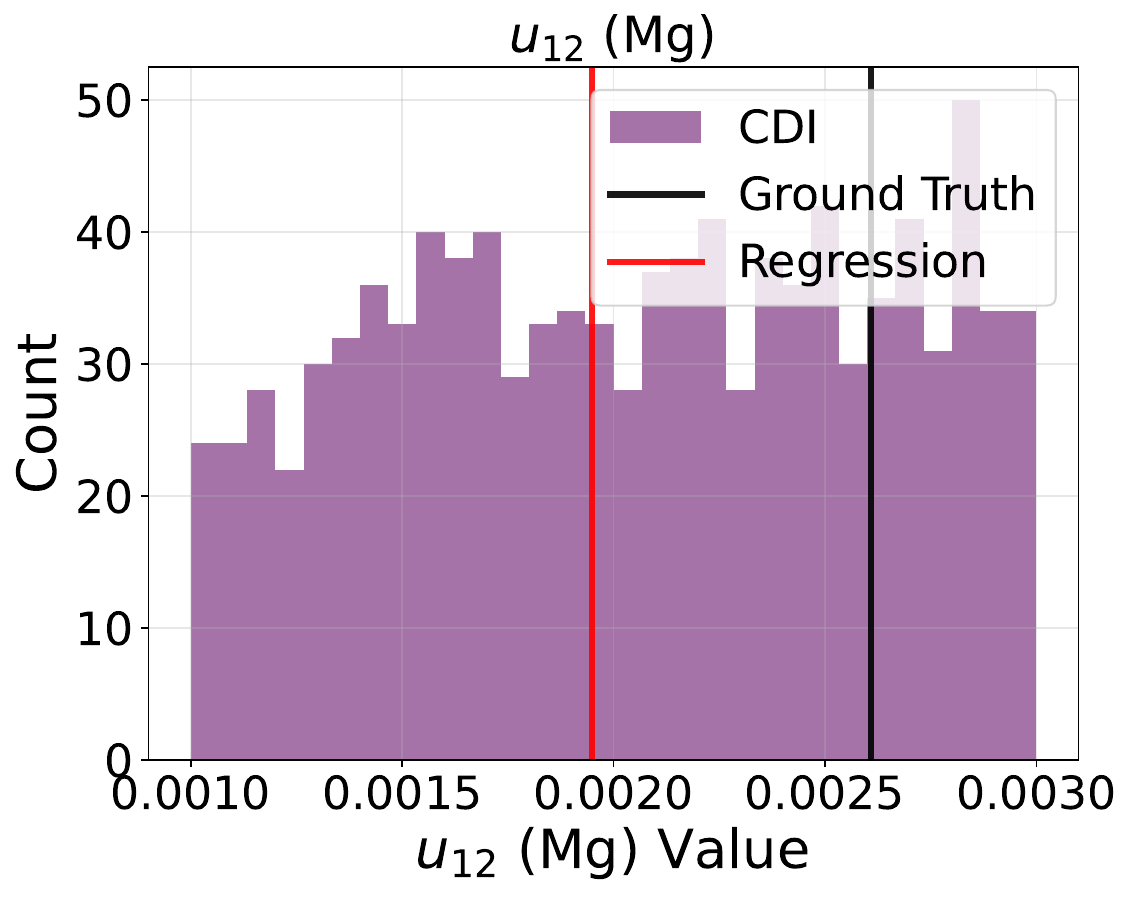} \\
        \includegraphics[width=0.33\textwidth]{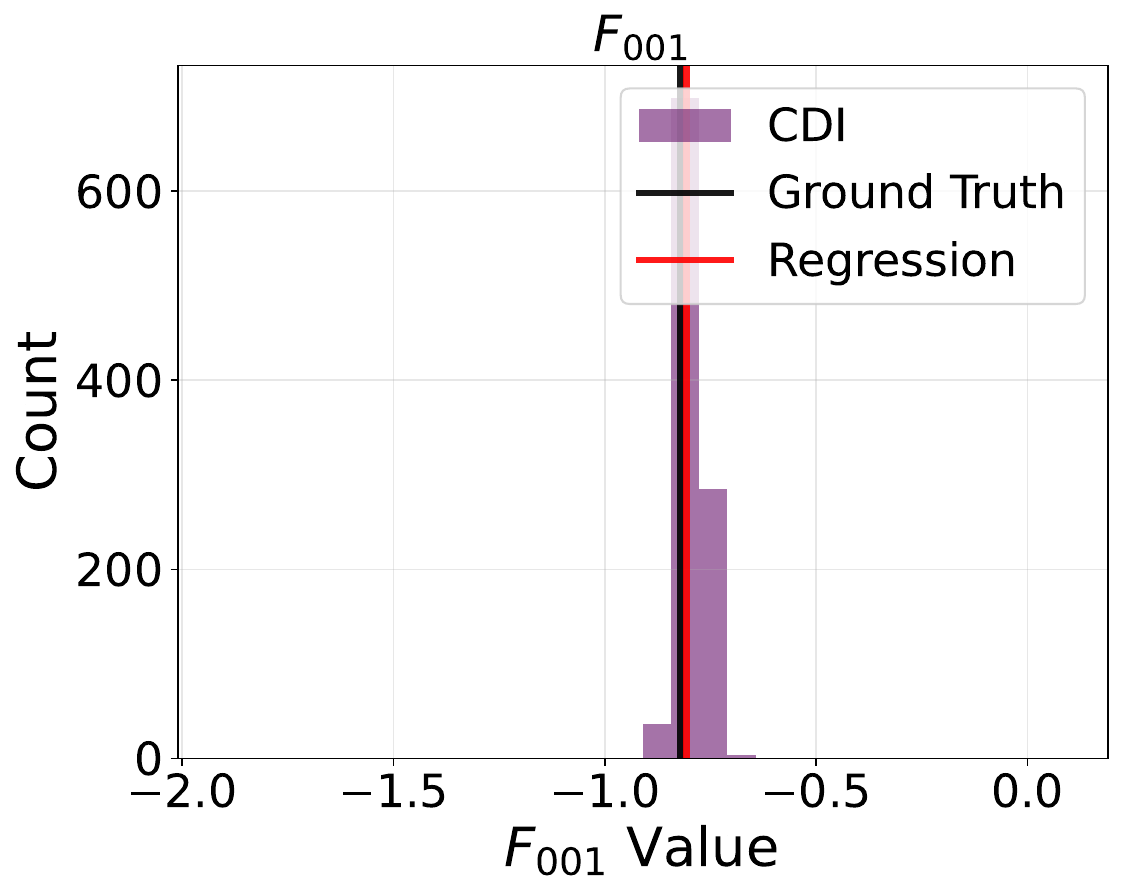} &
        \includegraphics[width=0.33\textwidth]{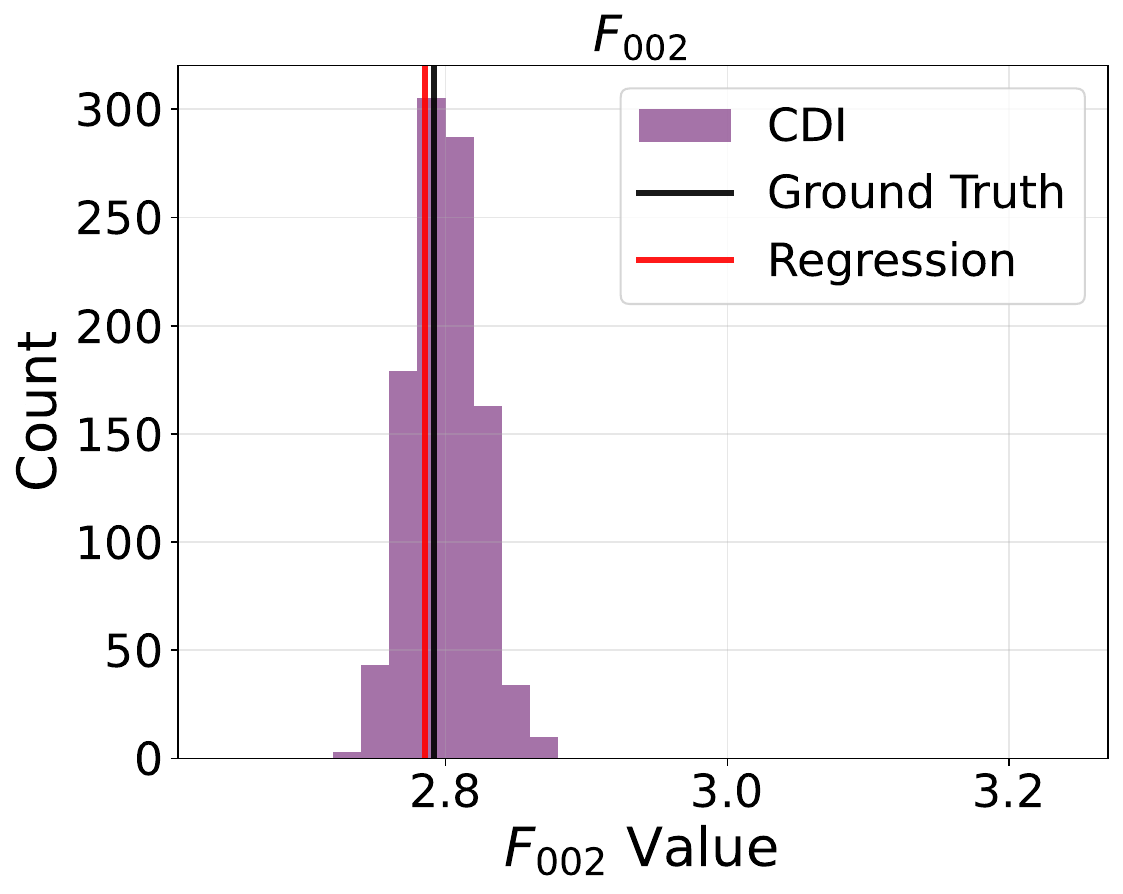} &
        \includegraphics[width=0.33\textwidth]{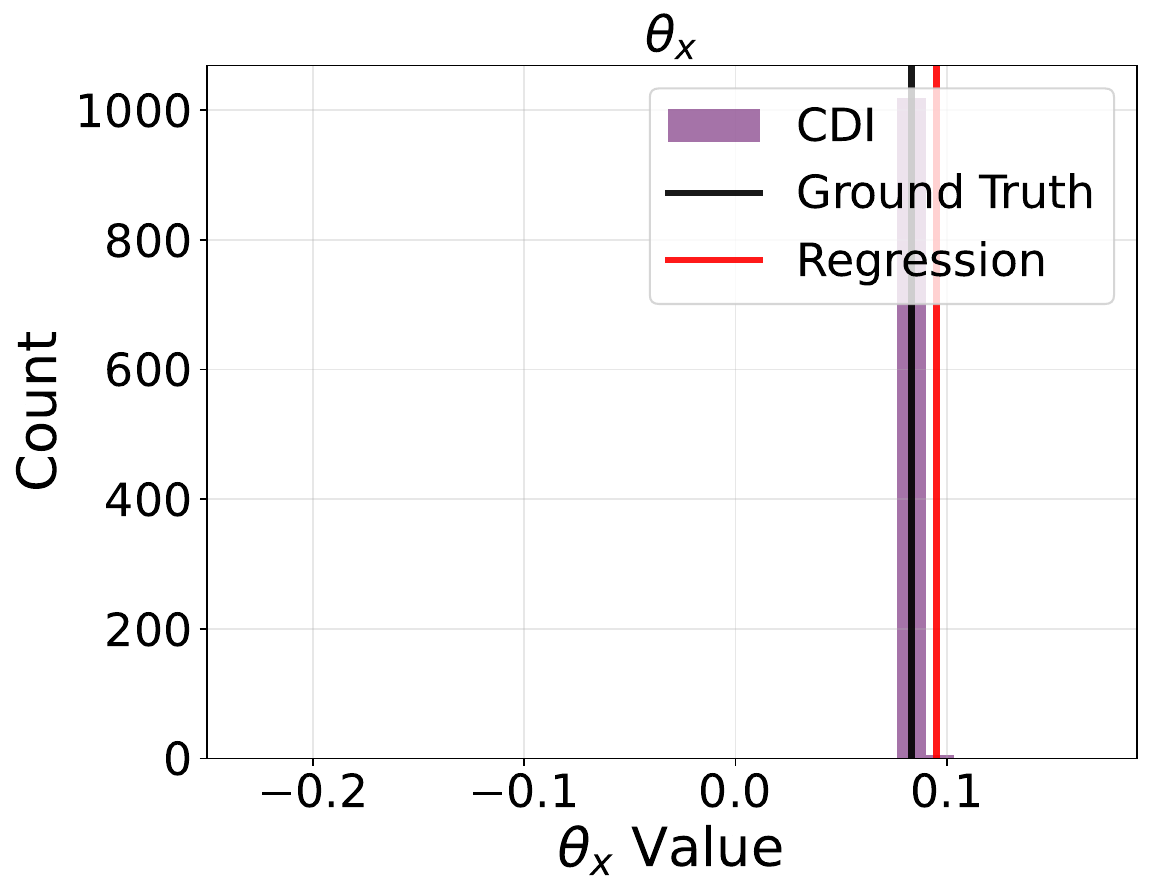}
    \end{tabular}
    \caption*{(b) Predicted Parameter Distributions (CDI vs Regression)}
\end{minipage}
\caption{Comparison of regression and CDI predictions for a representative CBED pattern. (a) Input diffraction pattern. (b) Posterior distributions for six representative parameters: purple histograms show CDI predictions from 1024 samples, black vertical lines indicate ground truth values, red vertical lines show regression point predictions.}
\label{fig:example_distributions}
\end{figure}

Figure~\ref{fig:example_distributions} compares CDI posterior distributions against regression point predictions for six representative parameters from a single CBED pattern. The predictions reveal substantial variation in parameter identifiability across different physical quantities.

Structure factors and sample geometry parameters demonstrate sharp, well-constrained posteriors. $F_{001}$ (bottom left) exhibits an extremely narrow distribution concentrated precisely at the ground truth value, indicating high confidence in the prediction. Similarly, the tilt angle $\theta_x$ (bottom right) shows a sharp posterior accurately capturing the true parameter value. Both CDI distributions and regression predictions align closely with ground truth for these well-constrained parameters, confirming that sufficient observational information exists to determine them precisely.

In contrast, Debye-Waller factors display varying degrees of uncertainty. $u_{11}$ (top left) shows a reasonably narrow distribution centered near ground truth, suggesting moderate constraint from the diffraction data.
However, $u_{12}$ (top right) exhibits a broad, nearly uniform distribution spanning much of the parameter range.
While this distribution contains the true value, its width indicates that the CBED pattern provides little information to constrain this parameter tightly.
The regression prediction for this parameter simply reproduces the mean of the training distribution. Likewise, regression estimates are biased to the mean for the other Debye-Waller factors (top row).

These examples illustrate CDI's key advantage: the posterior distributions automatically reflect the information content available in each observation. Well-constrained parameters yield sharp, confident distributions, while poorly constrained parameters produce appropriately broad posteriors that acknowledge uncertainty. Regression, by contrast, produces point estimates without indication of reliability, potentially conveying false confidence for ambiguous parameters.

\section{Discussion}

Our results demonstrate successful extension of the CDI framework from 1D temporal signal processing to 2D spatial pattern conditioning through a novel cross-modal transformer architecture.
This represents the first application of diffusion-based inverse problem solvers to high-dimensional image observations, establishing that parameter-space diffusion can effectively handle the substantially increased dimensionality and spatial structure of 2D data.
The architectural innovations—joint tokenization of heterogeneous data types, bidirectional cross-attention between spatial observations and parameter representations, and implicit positional encoding through hierarchical convolutional features—enable effective conditioning on image data while preserving the probabilistic inference capabilities of the original framework.
This validates that the core principle of parameter-space diffusion generalizes across data modalities when paired with appropriate architectural designs for observation encoding and cross-modal feature fusion.

Beyond demonstrating architectural flexibility, this work validates that CDI's uncertainty quantification capabilities translate effectively to the image-based CBED domain. The calibration analysis demonstrates well-aligned predicted and empirical confidence levels across most parameters, while coverage-sharpness analysis shows efficient posterior predictions for well-constrained quantities. Moreover, the behavior of CDI posterior distributions aligns with physical intuition: parameters with strong influence on diffraction patterns (geometry, structure factors) have well-constrained distributions, while parameters with weak influence (thermal vibrations parametrized by Debye-Waller factors) produce broader posteriors reflecting genuine observational ambiguity.

The contrast with regression baselines highlights CDI's practical value for scientific applications. Regression provides point estimates that can appear accurate when evaluated on aggregate metrics, yet these predictions mask underlying uncertainty structure. For poorly constrained parameters, regression defaults to training distribution means regardless of specific observational information, conveying false confidence where genuine ambiguity exists. CDI's distributional predictions provide faithful uncertainty representation, allowing one to distinguish between parameters determined precisely from data versus those requiring additional experimental constraints or prior knowledge.

\section{Conclusions}

We presented the first extension of the Conditional Diffusion Model-based Inverse Problem Solver (CDI) framework to image-based inverse problems, demonstrating its application to CBED parameter inference in materials characterization. Our work makes three key contributions: (1) a hybrid CNN-transformer architecture that adapts CDI from 1D temporal signals to 2D spatial patterns, (2) comprehensive uncertainty quantification evaluation demonstrating well-calibrated posterior distributions, and (3) validation that CDI's probabilistic predictions correctly reflect parameter-dependent measurement precision.

The results establish that CDI provides meaningful uncertainty quantification for physics inverse problems. Posterior distributions automatically adapt to parameter identifiability -- sharp and confident for well-constrained quantities like sample geometry and structure factors, appropriately broad for weakly-constrained parameters like Debye-Waller factors. This behavior aligns with physical intuition and provides essential information that regression baselines cannot capture. 

This work demonstrates CDI's potential as a flexible framework for probabilistic parameter inference across scientific domains. The straightforward architectural adaptation and strong performance on CBED data suggest promising directions for applying CDI to other materials characterization modalities and physics inverse problems requiring uncertainty-aware parameter estimation.
\subsubsection*{Acknowledgments}

This work was supported by the U.S. Department of Energy, Office of Basic Energy Science, Division of Materials Science and Engineering, under Contract No. DE-SC0012704.

\bibliographystyle{plainnat}
\bibliography{refs}

\section{Appendices}
\appendix

\section{CBED Simulation Details}
\label{sec:app_simulation}

CBED patterns are generated through physics-based forward simulations using the Bloch wave method. 
Given a set of material parameters (crystal structure, Debye-Waller factors describing thermal vibrations, valence electron distributions) and experimental conditions (sample thickness, and crystal orientation), the forward model calculates the expected diffraction pattern through quantum mechanical treatment of electron-crystal interactions.

This appendix summarizes the key equations and parameter definitions used in our simulation pipeline.
Dataset generation involves randomly sampling these parameters from physically realistic ranges and computing the corresponding CBED patterns, creating paired (image, parameters) data for training our inverse models.

\subsection{Bloch Wave Formulation}

CBED patterns are simulated using the Bloch wave method for high-energy electron diffraction. The electron wavefunction inside the crystal is expanded as:
\begin{equation}
\Psi(\mathbf{r}) = \sum_j c_j \exp(2\pi i \mathbf{k}^j \cdot \mathbf{r}) \sum_{\mathbf{g}} C_{\mathbf{g}}^j \exp(2\pi i \mathbf{g} \cdot \mathbf{r})
\end{equation}
where $\mathbf{g} = h\mathbf{a}^* + k\mathbf{b}^* + l\mathbf{c}^*$ is a reciprocal lattice vector.

Following the standard Bloch wave formalism~\cite{zuo2017advanced}, and neglecting back-scattered waves, the dispersion equation becomes:
\begin{equation}
2 K S_g C_{\mathbf{g}}^j + \sum_{\mathbf{h}} U_{\mathbf{g}-\mathbf{h}} C_{\mathbf{h}}^j = 2K_n \gamma^j C_{\mathbf{g}}^j
\end{equation}
where $\mathbf{K}$ is the incident electron wavevector, $K_n = \mathbf{K} \cdot \mathbf{n}$ is its component along the beam direction (with $\mathbf{n}$ a unit vector antiparallel to the beam), 
$S_g$ is the excitation error (distance between reflection $\mathbf{g}$ and the Ewald sphere), $U_{\mathbf{g}}$ are Fourier components of the crystal potential, and $\gamma^j \mathbf{n} = \mathbf{k}^j - \mathbf{K}$.

This can be written in matrix form $\mathbf{A}C^j = 2K_n\gamma^j C^j$, where eigenvalues $\gamma^j$ and eigenvectors $C^j$ are obtained by solving the structure matrix $\mathbf{A}$.

The diffraction intensity at thickness $t$ would be given by:
\begin{equation}
I_g(t) = \left|\sum_j c_j C_g^j \exp(2\pi i \gamma^j t)\right|^2
\end{equation}

\subsection{Debye-Waller and Structure Factors in Electron Scattering}

The diagonal elements of the structure matrix $\mathbf{A}$ depend on the excitation errors $S_g$, which are calculated from the incident beam geometry:
\begin{equation}
2KS_g = -2\mathbf{K} \cdot \mathbf{g} - |\mathbf{g}|^2
\end{equation}

The off-diagonal elements $U_{\mathbf{g}}$ depend on electron structure factors:
\begin{equation}
U_{\mathbf{g}} = \frac{\gamma F_{\mathbf{g}}^e}{\pi V}, \quad F_{\mathbf{g}}^e = \sum_j f_j^e(g) T_j(g) \exp(-2\pi i \mathbf{g} \cdot \mathbf{r}_j)
\end{equation}
where $\gamma$ is the relativistic constant, $V$ is the unit cell volume, $f_j^e(g)$ is the electron scattering factor for atom $j$ at position $\mathbf{r}_j$, and $T_j(g)$ is the temperature factor that accounts for thermal vibrations via anisotropic Debye-Waller factors (DWFs):
\begin{equation}
T_j(g) = \exp[-2\pi^2(u_{11}h^2 + u_{22}k^2 + u_{33}l^2 + 2u_{12}hk + 2u_{13}hl + 2u_{23}kl)]
\end{equation}
where $u_{11}, u_{22}, u_{33}, u_{12}, u_{13}, u_{23}$ are the six components of the DWF tensor.
For \chemfig{Mg B_2} (space group P6/mmm), Mg occupies the 6/mmm site at (0,0,0) and B occupies the -6m2 site at ($\frac{1}{3}$,$\frac{2}{3}$,$\frac{1}{2}$) and ($\frac{2}{3}$,$\frac{1}{3}$,$\frac{1}{2}$).
Point group symmetry imposes constraints: $u_{22} = u_{11}$ and $u_{13} = u_{23} = 0$ for both atomic sites. 
Our simulation framework includes DWF parameters $u_{11}$, $u_{33}$, and $u_{12}$ for each atom type.

\subsection{Valence Electron Corrections}

The independent atom model (IAM) calculates electron scattering factors $f_j^e(g)$ using tabulated values for neutral, isolated atoms~\cite{doyle1968relativistic}. However, valence electrons redistribute when atoms form chemical bonds in crystals, significantly altering scattering factors at low angles and causing systematic discrepancies between IAM-based simulations and experimental CBED patterns.

Quantitative accuracy requires accounting for bonding effects. In this work, we use the IAM-SF (Independent Atom Model with Structure Factor refinement) approach: a small number of low-order structure factors sensitive to valence electron distributions are treated as free parameters, while higher-order structure factors continue using IAM values. For \chemfig{Mg B_2}, we refine four low-order structure factors: $F_{001}$, $F_{100}$, $F_{101}$, and $F_{002}$.

\subsection{Dataset Generation}

We generate synthetic [010] CBED patterns by randomly sampling 13 parameters from physically realistic ranges:
\begin{itemize}
    \item Crystal DWFs: Mg ($u_{11}$, $u_{33}$, $u_{12}$) and B ($u_{11}$, $u_{33}$, $u_{12}$) -- 6 parameters
    \item Structure factors: $F_{001}$, $F_{100}$, $F_{101}$, $F_{002}$ -- 4 parameters
    \item Beam geometry: tilt angles ($\theta_x$, $\theta_y$) and sample thickness ($t$) -- 3 parameters  
\end{itemize}

Each parameter value is sampled uniformly. Table~\ref{tab:cbed_params} summarizes the parameter ranges.

\begin{table}[htb]
\centering
\begin{tabular}{|l|c|c|c|}
\hline
\textbf{Parameter} & \textbf{Mean} & \textbf{Range ($\pm$)} & \textbf{Units} \\
\hline
Mg $u_{11}$ & 48 & 25 & pm$^2$ \\
Mg $u_{33}$ & 56 & 25 & pm$^2$ \\
Mg $u_{12}$ & 20 & 10 & pm$^2$ \\
B $u_{11}$ & 47 & 25 & pm$^2$ \\
B $u_{33}$ & 51 & 25 & pm$^2$ \\
B $u_{12}$ & 20 & 10 & pm$^2$ \\
\hline
$F_{001}$ & -0.91 & 1.0 & \AA \\
$F_{100}$ & 0.63 & 0.6 & \AA \\
$F_{101}$ & 2.69 & 0.4 & \AA \\
$F_{002}$ & 2.94 & 0.3 & \AA \\
\hline
$\theta_x$ & -0.03 & 0.2 & deg \\
$\theta_y$ & 0.15 & 0.2 & deg \\
Thickness & 2800 & 1000 & \AA \\
\hline
\end{tabular}
\caption{Parameter ranges for CBED pattern simulation}
\label{tab:cbed_params}
\end{table}

\begin{figure*}[htb]
    \centering
    \begin{subfigure}[b]{0.45\textwidth}
        \centering
        \includegraphics[width=\textwidth]{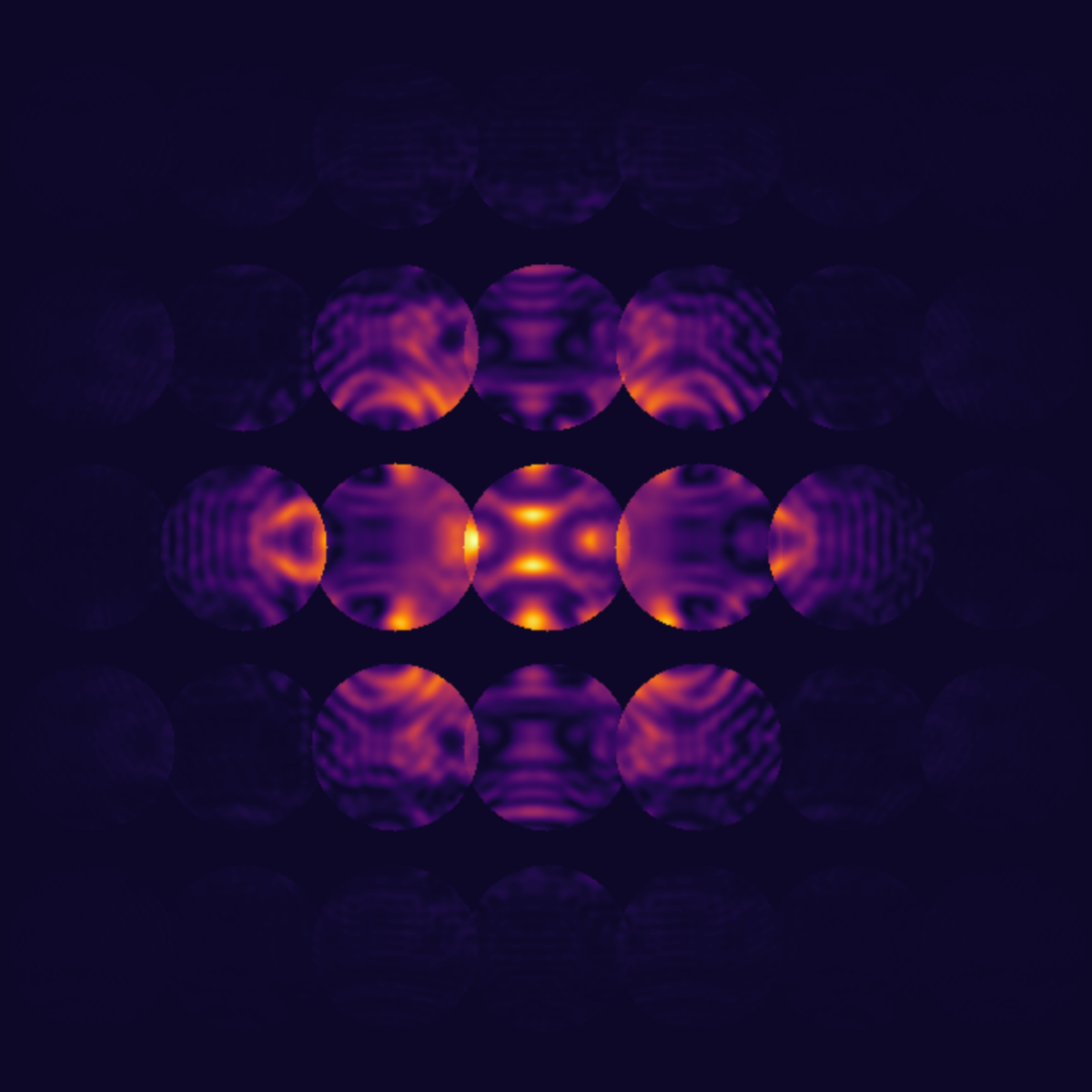}
    \end{subfigure}
    \caption{Representative simulated CBED pattern for MgB$_2$ generated via Bloch wave method.
    The $640 \times 640$ pixel image captures the central $3 \times 3$ array of primary diffraction discs (bright features) along with surrounding higher-order reflections.
    Intensity modulations within each disc encode information about crystal structure, sample thickness, beam orientation, and valence electron distributions.
    }
    \label{fig:cbed_full_example}
\end{figure*}

Each simulated CBED pattern is rendered as a $640 \times 640$ pixel image capturing the central region of the diffraction pattern.
The resulting images contain the primary $3 \times 3$ array of diffraction discs along with surrounding higher-order reflections, as shown in Figure~\ref{fig:cbed_full_example}. The intensity modulations within and across diffraction discs encode the underlying parameter values -- variations in sample thickness and orientation affect overall intensity oscillations, Debye-Waller factors influence high-angle intensity falloff, and structure factors modulate fine intensity details within discs. This physics-based forward model generates the training data for our inverse problem approach.

For model training, these full simulated patterns undergo preprocessing and cropping to match realistic experimental field-of-view constraints, as detailed in Appendix~\ref{sec:app_training_data}.

\section{Training Details}
\label{sec:app_training}

\subsection{Data Preprocessing and Augmentation}
\label{sec:app_training_data}

Input CBED patterns ($640 \times 640$ pixels) undergo preprocessing before model training and evaluation.
During training, we apply data augmentation to simulate experimental variability and imperfect data acquisition conditions: random rotation ($\pm 15^\circ$) with bilinear interpolation to account for arbitrary sample orientations during mounting, followed by center cropping to $400 \times 400$ pixels to match typical experimental field-of-view, and finally random resized cropping to $320 \times 320$ pixels with scale factors between $0.6 - 1.0$ to simulate variations in microscope magnification and framing. 

During evaluation, patterns are simply center-cropped to $320 \times 320$ pixels to simulate typical experimental field-of-view. All models (regression baselines and CDI) use identical preprocessing pipelines for fair comparison.

\subsection{Regression Baseline Training}

All regression baselines (ResNet, EfficientNet) are trained using identical configurations for fair comparison. Models are trained for 500 epochs with batch size 32 using the AdamW optimizer with learning rate $10^{-4}$ and zero weight decay. Training uses 1000 gradient steps per epoch on the training set. Learning rate scheduling uses cosine annealing with $T_{\text{max}} = 500$ epochs.

Target parameters are normalized before training using the mean values and ranges listed in Table~\ref{tab:cbed_params}. The loss function is mean squared error (MSE) computed on normalized parameter predictions.

\subsection{CDI Architecture}
\label{sec:app_cdi_arch}

The CDI model uses a hybrid vision-transformer architecture for image-conditioned parameter diffusion.
The vision encoder is a ResNet-34 backbone followed by a downsampling convolutional layer (512 features, $4 \times 4$ kernel, stride 2, padding 1) and an additional ResNet block, producing 512-dimensional features at $5 \times 5$ spatial resolution. These spatial features are flattened and projected through a linear layer to create 25 vision tokens of dimension 256.

Each of the 13 target parameters is encoded as an individual token through dedicated linear projections. The diffusion timestep $t$ is encoded as a single token via linear projection. The transformer encoder processes the concatenated sequence of 13 parameter tokens, 1 time token, and 25 vision tokens (39 tokens total) through 8 layers with 256-dimensional embeddings, 8 attention heads, and 1024-dimensional feedforward layers using pre-normalization. 

Individual linear decoders project each output parameter token back to scalar parameter predictions.

\subsection{CDI Training}

The diffusion model is trained for 1000 epochs with batch size 64 using the AdamW optimizer with learning rate $2 \times 10^{-4}$ and zero weight decay. Training uses 1000 gradient steps per epoch. No learning rate scheduling is applied. An exponential moving average (EMA) of model weights is maintained with momentum 0.9999.

We use a linear noise schedule with $T=800$ diffusion timesteps, where $\beta_1 = 2.5 \times 10^{-5}$ and $\beta_T = 0.02$. Target parameters are normalized identically to regression baselines using the mean values and ranges from Table~\ref{tab:cbed_params}.

\end{document}